\documentclass[runningheads]{llncs}

\usepackage{eccv}

\usepackage{eccvabbrv}

\usepackage{graphicx}
\usepackage{booktabs}
\usepackage{multirow}
\usepackage[table]{xcolor}
\usepackage{enumitem}
\usepackage{mathtools}
\usepackage{ragged2e}
\usepackage{hyperref}
\usepackage[accsupp]{axessibility}  % Improves PDF readability for those with disabilities.
\definecolor{highlightblue}{RGB}{235, 245, 255}

\usepackage{hyperref}

\usepackage{orcidlink}

\begin{document}

\title{OTCache: Optimal Transport for Geometry-Aware Caching in Diffusion Models} 

\titlerunning{OTCache}

% % TODO FINAL: Replace with your author list. 
% % Include the authors' OCRID for the camera-ready version, if at all possible.
% Second Author\inst{2,3}\orcidlink{1111-2222-3333-4444} \and
% Third Author\inst{3}\orcidlink{2222--3333-4444-5555}}

% % TODO FINAL: Replace with an abbreviated list of authors.

% % TODO FINAL: Replace with your institution list.
% Springer Heidelberg, Tiergartenstr.~17, 69121 Heidelberg, Germany

\author{
Huanlin Gao\inst{1,2}$^\dagger$\orcidlink{0009-0006-3742-6023} \and
Fang Zhao\inst{1,2}$^\dagger$\orcidlink{0009-0003-3465-3050} \and
Qiang Hui\inst{1,2}\orcidlink{0000-0002-3674-092X} \and
Fuyuan Shi\inst{1,2} \and
Shaoan Zhao\inst{1,2} \and
Yantao Li\inst{1,2,3} \and
Chao Tan\inst{1,2}\orcidlink{0009-0008-4957-6939} \and
Ting Lu\inst{1,2} \and
Yuren You\inst{2}\orcidlink{0009-0004-0168-9265} \and
Kai Wang\inst{1,2}$^*$\orcidlink{0000-0002-1171-0281} \and
Shiguo Lian\inst{1,2}$^*$\orcidlink{0000-0003-4308-7049}
}

\authorrunning{H. Gao et al.}

\institute{
Data Science \& Artificial Intelligence Research Institute, China Unicom
\and
Unicom Data Intelligence, China Unicom
\and
National Key Laboratory for Novel Software Technology, Nanjing University
\\
\email{\{gaohl51,zhaof50,wangk115,liansg\}@chinaunicom.cn}
\\
$^\dagger$Equal contribution. 
$^*$Corresponding authors.
% \\
% % Code: \url{https://github.com/UnicomAI/OTCache}
}

\maketitle

\begin{abstract}
We propose \textbf{OTCache}, a training-free framework for accelerating diffusion sampling via caching schedule prediction. Existing graph-based caching methods reduce redundant computation by optimizing shortest-path objectives, but rely on an additive independence assumption, which often breaks down in the low NFE regime. To address this issue, OTCache models caching schedules across inference budgets as a smooth evolution in policy space, inspired by Optimal Transport (OT). The framework consists of three stages: (1) obtaining a high-fidelity \textbf{reference schedule} using a graph-based caching method under a conservative budget; (2) performing a lightweight \textbf{anchor search} under an extreme low-budget setting via Optuna optimization with an end-to-end perceptual objective; and (3) predicting schedules for target budgets via \textbf{quantile interpolation} between the reference and anchor policies using continuous warping representations. Experiments on FLUX.1 [dev], Qwen-Image, and HunyuanVideo show that OTCache achieves $4.5\times$, $4.7\times$, and $3.66\times$ acceleration, respectively, while consistently improving generation fidelity over state-of-the-art caching baselines. This work provides a new perspective on accelerating diffusion models through Optimal-Transport-inspired schedule modeling. \textbf{Code:} \url{https://github.com/UnicomAI/OTCache}
\keywords{Diffusion Models \and Graph-based Caching \and Optimal Transport}
\end{abstract}

\section{Introduction}

Flow Matching (FM) \cite{lipman2022flow, albergo2022building} has emerged as a cornerstone of modern generative modeling, driving significant breakthroughs in high-fidelity synthesis across image \cite{flux2024}, video \cite{kong2024hunyuanvideo, Open-Sora}, and multi-modal tasks \cite{hung2024tangofluxsuperfastfaithful}. By modeling continuous transport paths via instantaneous velocity fields, FM provides a mathematically principled and effective paradigm for sampling. However, in commercial-scale models such as FLUX.1 \cite{flux2024} and HunyuanVideo \cite{kong2024hunyuanvideo}, the massive parameter scale of transformer-based denoisers coupled with the prolonged iterative sampling process leads to staggering computational overhead and memory footprints. This \textbf{computation-intensive} nature creates a formidable deployment barrier, significantly hindering their applicability in interactive or resource-constrained scenarios.

To alleviate this burden, established acceleration strategies such as distillation \cite{salimans2022progressive, sauer2024adversarial}, pruning \cite{han2015deep}, and quantization \cite{li2023q} have been widely explored. Nevertheless, most of these methods impose a heavy ``adoption tax''---they typically necessitate intensive retraining on large-scale datasets, complex architectural modifications, or sophisticated engineering pipelines, which limits their flexibility for rapid deployment. In contrast, \textbf{caching techniques} \cite{ma2023deepcache, wimbauer2023cache} have emerged as a compelling training-free alternative. 

\begin{figure*}[t]
  \centering
  \includegraphics[width=0.85\linewidth]{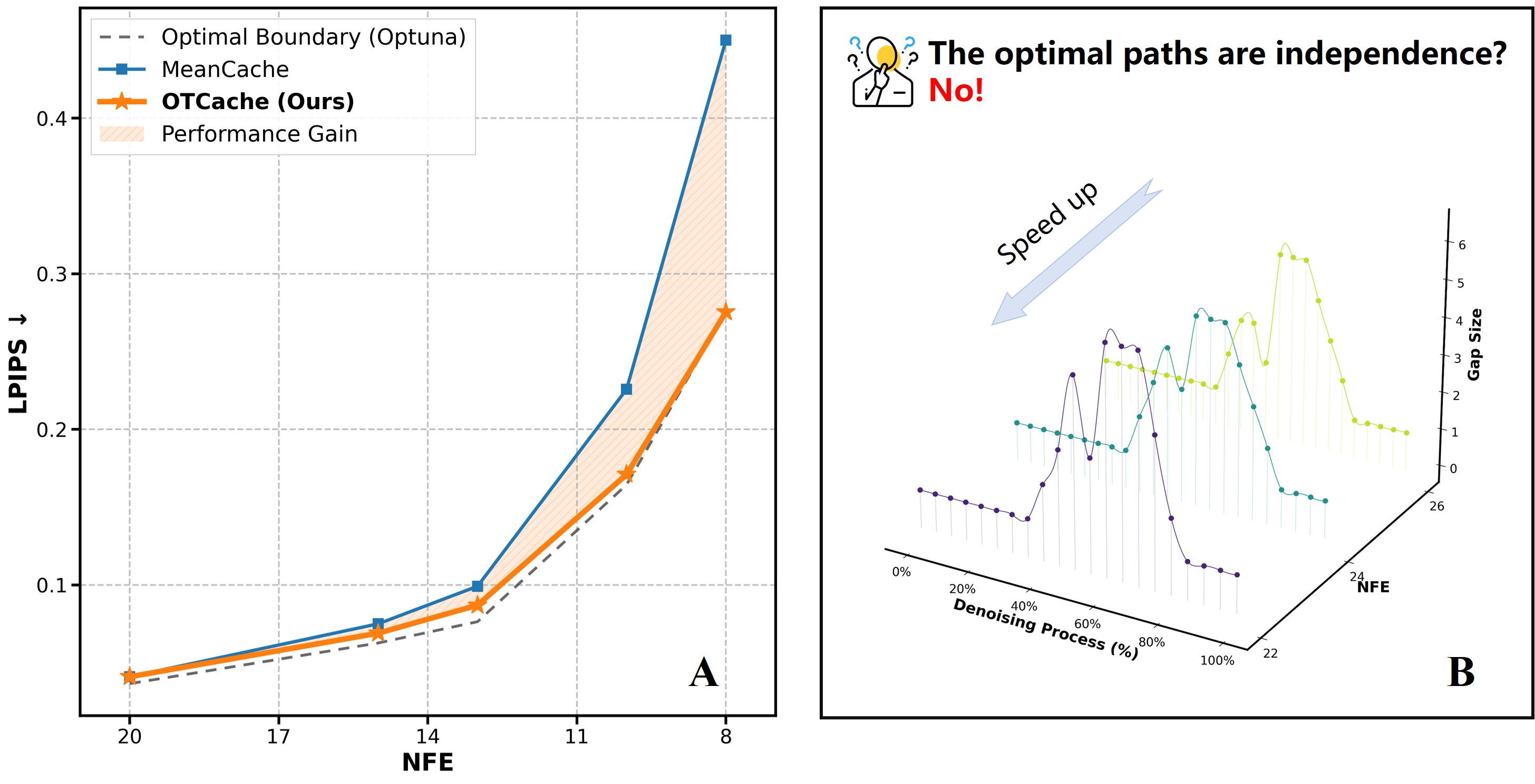}
  \vspace{-10pt}
  \caption{
    \textbf{Performance gain and structural insights on FLUX.1 [dev].} 
    \textbf{(A) Failure of additive surrogates:} MeanCache deviates from the optimal boundary in the ultra-low NFE regime, reflecting the limitations of local error aggregation. OTCache recovers this fidelity loss, significantly narrowing the gap to the search-based optimum. 
    \textbf{(B) Structural regularity of optimal paths:} Optimal schedules across different NFE (budgets) demonstrate strong structural relationships rather than independent patterns, suggesting a shared underlying policy trajectory that OTCache exploits via geometric interpolation.
  }
  \vspace{-5mm}
  \label{fig:motivation}
\end{figure*}

Recent advancements such as MeanCache~\cite{gao2026meancache} formulate cache scheduling as a constrained shortest-path problem, enabling efficient, training-free planning under a fixed NFE budget. 
However, this paradigm relies on an \emph{additive independence} assumption, approximating end-to-end degradation as a sum of local edge errors. We find that this surrogate becomes increasingly unreliable in the low-NFE regime. As shown in Fig.~\ref{fig:motivation}A, when NFE decreases from 20 to 8, MeanCache still exhibits a clear gap to the search-based optimum (Optuna), indicating substantial room for improvement in LPIPS.

More fundamentally, Fig.~\ref{fig:motivation}B reveals an overlooked structural regularity: the gap profiles (first-order differences) of optimal schedules across budgets are \textbf{not} unrelated; instead, their temporal structures evolve smoothly as the NFE budget varies. This observation motivates a shift in perspective, rather than independently solving a discrete shortest-path problem for each budget, we model budget-conditioned schedule evolution and exploit structural relationships across budgets.

Based on this insight, we propose \textbf{OTCache}, a training-free framework that predicts schedules for Low-NFEs via optimal transport interpolation. By anchoring prediction with a reliable high-budget reference and a low-budget anchor, and interpolating between them in Wasserstein space, OTCache generates robust schedules that remain stable under Low-NFE acceleration. The main contributions of this work are summarized as follows:
\begin{itemize}
    \item \textbf{Rethinking Cache Scheduling under Low NFE.} 
    We identify two fundamental limitations of existing graph-based schedulers: (i) surrogate misalignment caused by additive shortest-path objectives under nonlinear error propagation, and (ii) the independent-budget assumption that ignores structural relationships across optimal schedules.

    \item \textbf{Budget-Conditioned Schedule Evolution.}
    We introduce a three-stage acceleration framework that treats cache schedules as probability measures and models their evolution across NFEs as a smooth trajectory in policy space. Leveraging the geometric structure of Optimal Transport (OT), we obtain target-budget schedules through quantile interpolation between two reliable endpoints. 

    \item \textbf{Outstanding Performance.}
   Experiments on FLUX.1, Qwen-Image, and HunyuanVideo demonstrate that OTCache achieves 4.5$\times$, 4.7$\times$, and 3.66$\times$ acceleration, respectively. In particular, on Qwen-Image, OTCache achieves an LPIPS of \textbf{0.171}.
\end{itemize}

\section{Related Work}

\subsection{Diffusion Model Acceleration.} 
The remarkable success of generative models \cite{song2020ddim,song2020score} across diverse modalities is significantly attributed to the continuous advancements in sampling speed. Early efforts primarily focused on optimizing the iterative denoising process through principled numerical SDE/ODE solvers \cite{song2020score,jolicoeur2021gotta,chen2025optimizing}, such as DDIM \cite{song2020ddim}, EDM \cite{karras2022elucidating}, and DPM-Solver \cite{lu2022dpm}, which aim to maintain high synthesis quality with fewer discretization steps. To further compress inference trajectories, knowledge distillation \cite{hinton2015distilling} has been extensively explored to map multi-step denoising into compact few-step or even single-step regimes, exemplified by Progressive Distillation \cite{salimans2022progressive} and Consistency Models \cite{song2023consistency, sauer2024adversarial,wang2025target}. Complementary to these, orthogonal strategies—including quantization \cite{li2023q, shang2023post}, pruning \cite{ma2023llm}, and system-level parallelization frameworks \cite{zhao2024dsp, chen2024asyncdiff}—have been investigated to enhance raw hardware throughput. More recently, the emergence of Flow Matching (FM) \cite{lipman2022flow, albergo2022building} has introduced a new paradigm that learns deterministic velocity fields, inherently possessing the potential for high-fidelity generation with minimal sampling steps. Nevertheless, most existing methods still require heavy computation, large-scale data, or complex engineering, limiting their practical adoption in resource-constrained scenarios.

\subsection{Cache in Diffusion Models.} 
As a training-free acceleration paradigm, caching strategies \cite{wimbauer2023cache, ma2024learning} have gained prominence by reusing intermediate representations to bypass redundant computation. Early methods such as DeepCache \cite{ma2023deepcache} introduced architecture-specific feature reuse for UNet backbones, while T-GATE \cite{zhang2024cross} and $\Delta$-DiT \cite{chen2024delta} extended this idea to Transformer-based architectures \cite{peebles2023dit}. For large-scale video generation, PAB \cite{zhao2024real} and TeaCache \cite{DBLP:journals/corr/abs-2411-19108} exploit temporal correlations and error thresholds to trigger feature reuse. Recent studies further explore fine-grained caching criteria, including adaptive reuse for video diffusion transformers \cite{kahatapitiya2025adaptive}, profiling-based cache selection \cite{ma2025model}, frequency-aware caching \cite{liu2025freqca}, speculative feature caching \cite{liu2025speca}, and search-based policy discovery \cite{aggarwal2025evolutionary}. The field has also evolved towards graph-based caching; LeMiCa \cite{gao2025lemica} abstracts video synthesis into a Directed Acyclic Graph (DAG) for global scheduling, while MeanCache \cite{gao2026meancache} draws inspiration from MeanFlow \cite{geng2025mean} and reformulates caching from an instantaneous velocity view to an average velocity perspective, stabilizing the sampling trajectory. Despite these advances, existing methods typically optimize schedules within a fixed budget or rely on local surrogate criteria. How to obtain accurate \emph{optimal} caching policies in the extreme high-acceleration regime, namely the low-NFE setting, remains an open problem.

\section{Methodology}

\subsection{Preliminaries}

\subsubsection{Flow Matching.}
Flow Matching \cite{lipman2022flow,albergo2022building} and Rectified Flow \cite{liu2022flow} introduce a new paradigm for diffusion-based generative modeling by constructing continuous transport paths between a noise distribution $\pi_1$ and a data distribution $\pi_0$. By defining a probability path via linear interpolation:
\begin{equation}
    x_t = (1-t)x_0 + t x_1, \quad t \in [0, 1],
\end{equation}
these methods aim to learn a straight-line trajectory in the state space. Since the clean data $x_0$ is unknown during the denoising (generation) process, a learned velocity field $v_\theta(x_t, t)$ is employed to approximate the direction $(x_0 - x_1)$, thereby constructing a neural ODE model:
\begin{equation}
    d\hat{x}_t = v_\theta(x_t, t) dt
\end{equation}
Numerical solvers discretize this ODE into $N$ steps to recover the data distribution. The total computational cost is primarily determined by the Number of Function Evaluations (NFE), making efficient step-wise scheduling critical for inference acceleration.

\vspace{-5pt}

\subsubsection{Graph-based Cache Scheduling}
\label{sec:Graph_based_Cache}
Feature caching has emerged as a predominant training-free paradigm to accelerate inference by reusing intermediate states across adjacent timesteps. Recent state-of-the-art methods~\cite{gao2025lemica,gao2026meancache} formulate this as a constrained shortest-path problem on a directed multigraph $\mathcal{G}=(\mathcal{V}, \mathcal{E})$. In this formulation, each node $v \in \mathcal{V}$ corresponds to a discrete timestep, and an edge $e = (t \to s)$ represents a caching transition where the velocity field at $s$ is estimated using cached features from $t$. The fidelity loss is quantified by edge weights:
\begin{equation}
    \mathcal{L}(t \to s) = \| v(x_s, s) - \hat{v}(x_s, s; x_t) \|_p,
    \label{eq:edge_weight}
\end{equation}

where $\hat{v}$ denotes the cached estimator. The global scheduling problem aims to find an optimal path $\pi^\star$ that minimizes the cumulative error:
\begin{equation}
    \pi^\star = \arg\min_{\pi \in \mathcal{P}} \sum_{e \in \pi} \mathcal{L}(e)^{\gamma}, \quad \text{s.t. } |\pi| \leq \mathcal{B},
    \label{eq:shortest_path}
\end{equation}
where $\mathcal{P}$ is the set of feasible paths and $\mathcal{B}$ is the computation budget.

\subsection{Rethink: Beyond Additive Shortest-Path Surrogates}
\label{sec:rethink}

\paragraph{Limitation.}
Graph-based schedulers formulate caching as a constrained shortest-path problem (Eq.~\ref{eq:shortest_path}). 
In this formulation, the caching cost between any two timesteps $(t,s)$ is first estimated independently as an edge weight $\mathcal{L}(e)$, and the optimal schedule is obtained by minimizing the sum of these edge costs. 
This implicitly assumes \emph{additive independence}: the end-to-end degradation of a schedule can be approximated by aggregating independently estimated edge losses.

This approximation is generally reliable in the \textbf{high-NFE regime}, where caching intervals are short and each decision has limited influence on subsequent steps. 
However, in the \textbf{low-NFE regime}, both the number of caching operations and the spacing between cached states increase. 
Consequently, the effect of earlier caching decisions propagates further along the trajectory and interacts with later ones, violating the additive independence assumption. 
As a result, the shortest-path surrogate becomes increasingly inaccurate for estimating the true path degradation. 
Consistent with this observation, Fig.~\ref{fig:motivation}A shows that the performance gap between MeanCache and the optimal boundary grows as NFE decreases.

\paragraph{Observation.}
Despite this limitation, we observe a clear structural regularity: near-optimal schedules across different NFE budgets are \emph{not independent}. 
As shown in Fig.~\ref{fig:motivation}B, the temporal gap patterns of optimal schedules evolve smoothly as the budget varies. 
This suggests that optimal schedules under different NFEs follow a structured evolution rather than forming unrelated solutions.

\paragraph{Core Insight.}
We hypothesize that optimal schedules under different NFE budgets correspond to different-resolution observations of a shared \emph{ideal trajectory} in policy space. Although the observation resolution (NFE) changes, the underlying trajectory remains fixed. Under this perspective, schedules across budgets form a smooth evolution on a policy manifold, which we interpret as a \emph{policy geodesic}. This view motivates predicting schedules through geometric interpolation in strategy space, which we instantiate using optimal transport in the following section.

\begin{figure*}[h]
  \centering
  \vspace{-2pt}
  \includegraphics[width=0.95\linewidth]{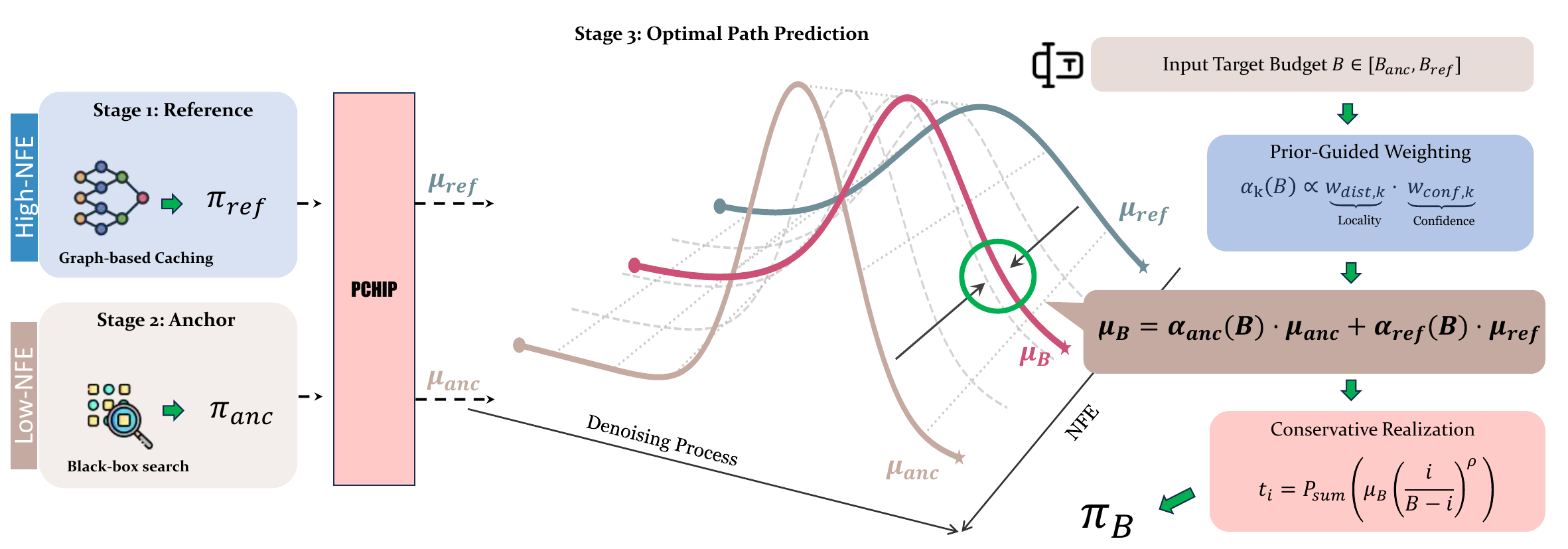}
  \vspace{-2pt}
    \caption{
    \textbf{Overview of OTCache}. Stage~1: use graph-based caching methods to obtain a reliable high-NFE(budget) schedule as the reference policy. Stage~2: perform black-box search to find a near-optimal low-NFE schedule as the anchor policy. Stage~3: inspired by optimal transport (OT), convert both endpoint schedules into continuous warping curves via PCHIP and apply quantile interpolation to predict the target-budget schedule $\pi_B$.
    }
  \vspace{-5mm}
  \label{fig:OTCache_overivew}
\end{figure*}

\subsection{OTCache}

Motivated by the observations in Sec.~\ref{sec:rethink}, we propose \textbf{OTCache}, a training-free, three-stage framework for accelerating diffusion sampling via caching schedule prediction, as illustrated in Fig.~\ref{fig:OTCache_overivew}.

\subsubsection{Stage 1: Reference}
\label{sec:stage1}

Empirically, we find that graph-based caching methods (e.g., MeanCache) may fail to recover the globally optimal schedule in the high-acceleration regime (low-NFE regime), where large integration steps can break the validity of additive shortest-path surrogates. In contrast, under a conservative budget, discretization errors are sufficiently mild such that the surrogate objective aligns well with end-to-end generation fidelity. Throughout this work, we adopt a conservative reference budget.

We obtain a reference schedule using a generic graph-based caching method under this conservative setting. Formally, we denote the resulting schedule as the reference policy:
\begin{equation}
\pi_{\mathrm{ref}}
\;\coloneqq\;
\mathcal{C}_{\mathrm{graph}}\!\left(B_{\mathrm{ref}}\right).
\label{eq:stage1_ref}
\end{equation}

Here $\mathcal{C}_{\mathrm{graph}}(\cdot)$ denotes a generic graph-based caching method; in our experiments, we instantiate it with MeanCache. We treat $\pi_{\mathrm{ref}}$ as a high-fidelity reference schedule in the conservative regime, which provides a structural prior for the subsequent budget-conditioned schedule prediction.

\subsubsection{Stage 2: Anchor}
\label{sec:stage2}

In the low-NFE regime, MeanCache can be unreliable because its shortest-path formulation relies on the \emph{additive independence} assumption, which may break down under large integration steps. As a result, the path minimizing the additive surrogate can be misaligned with true end-to-end generation fidelity. To obtain an accurate low-budget boundary condition for subsequent schedule prediction, we introduce an \emph{anchor} schedule by directly optimizing an end-to-end perceptual objective under an extreme budget $B_{\mathrm{anc}}$ via black-box search.

\paragraph{End-to-end objective.}
Let $x_0(\pi)$ denote the generated sample obtained by executing a candidate schedule $\pi$, and $x_0$ denote the full-budget reference output obtained once by running the original (non-cached) sampling path with all steps evaluated under the same prompt and seed. We define the anchor objective as the perceptual discrepancy between the two outputs:

\begin{equation}
\mathcal{J}(\pi)
\;\triangleq\;
\ell_{\mathrm{LPIPS}}\!\left(x_0(\pi),\, x_0\right).
\label{eq:stage2_e2e_obj}
\end{equation}

This objective is defined per prompt-seed pair. Unlike MeanCache, which optimizes a surrogate based on averaged velocity discrepancies, and LeMiCa, which optimizes latent-space deviations, we directly optimize a perceptual distance (LPIPS\cite{zhang2018unreasonable}) on the final generated image or video outputs, thereby capturing the fidelity gap induced by acceleration in an end-to-end manner.

\paragraph{Black-box search.}
To obtain an accurate anchor in the low budget regime, we directly minimize the end-to-end objective in Eq.~\eqref{eq:stage2_e2e_obj} using a lightweight black-box optimizer (Optuna + CMA-ES). We run the search for a fixed number of trials and apply early stopping to cap the cost (i.e., terminate if no improvement is observed for a predefined patience):
\begin{equation}
\pi_{\mathrm{anc}}
\;\coloneqq\;
\underset{\pi \in \mathcal{P}_{B_{\mathrm{anc}}}}{\arg\min}\;
\mathcal{J}(\pi),
\end{equation}
where $\mathcal{P}_{B_{\mathrm{anc}}}$ denotes the feasible set under budget $B_{\mathrm{anc}}$.

To improve sample efficiency, we adopt two practical choices. 
\textbf{(i) Warm start.} We initialize the search from the MeanCache schedule at the same budget (e.g., $B_{\mathrm{anc}}=8$), leveraging a strong prior and providing a conservative fallback when no improvement is found for a given prompt-seed pair.
\textbf{(ii) First-order parameterization.} We optimize the schedule in a gap (first-order difference) space rather than directly over discrete timesteps, which empirically yields a better-conditioned search landscape while preserving monotonic structure by construction.

\subsubsection{Stage 3: Optimal Path Prediction}
\label{sec:stage3}

To generalize scheduling patterns across arbitrary budgets, we model their evolution as a continuous trajectory in a functional space.

\paragraph{Continuous Representation via Monotonic Splines.}
We lift each discrete schedule $\pi_B=\{t_i\}_{i=0}^{B-1}$ to a continuous warping function over a normalized progress domain $u\in[0,1]$ by assigning uniformly spaced progress coordinates and fitting a shape-preserving Piecewise Cubic Hermite Interpolator (PCHIP). This yields a strictly monotone time-warping curve without spurious oscillations. The curve is then uniformly sampled at a fixed resolution to obtain an equal-dimensional representation. Applying this procedure to $\pi_{\mathrm{ref}}$ and $\pi_{\mathrm{anc}}$ produces the aligned continuous embeddings $(\mu_{\mathrm{ref}}, \mu_{\mathrm{anc}})$, enabling stable optimal-transport interpolation across budgets.

\paragraph{Log-conditioned Weighted Geodesic.}
To predict the schedule for a target budget $B \in [B_{\mathrm{anc}}, B_{\mathrm{ref}}]$, we model the transition between policies as a smooth path in the space of probability measures. In one dimension, the most natural way to interpolate between two distributions is through the \textbf{Wasserstein geodesic}. A key property of this approach is that the interpolation can be performed directly as a linear combination of the \textbf{quantile functions} (i.e., our continuous warping curves $\mu(u)$). 

Specifically, the predicted schedule $\mu_B(u)$ for target budget $B$ is constructed as:

\begin{equation}
    \mu_B(u) = \alpha_{\mathrm{anc}}(B) \cdot \mu_{\mathrm{anc}}(u) + \alpha_{\mathrm{ref}}(B) \cdot \mu_{\mathrm{ref}}(u),
\end{equation}

where the interpolation weights $\alpha_k(B)$ are defined by:

\begin{equation}
    \alpha_k(B) = \frac{w_{\text{dist}, k} \cdot w_{\text{conf}, k}}{\sum_{j \in \{\text{anc, ref}\}} w_{\text{dist}, j} \cdot w_{\text{conf}, j}}.
\end{equation}

Here, the terms $w_{\text{dist}, k} = (|B_k - B| + 1)^{-1}$ and $w_{\text{conf}, k} = \log B_k$ represent two intuitive priors: (i) \textbf{Locality}, which anchors the prediction to the nearest known budget; and (ii) \textbf{Confidence}, which assigns higher trust to the high-budget reference since dense schedules provide a more stable structural prior of the ODE trajectory. 

\paragraph{Conservative Realization.}
The discrete schedule $\pi_B = \{t_i\}_{i=0}^{B-1}$ is realized by sampling the continuous geodesic $\mu_B$ via a \textbf{power-law warping}. Motivated by the observation that early-stage caching requires more conservative updates due to high vector field volatility \cite{gao2025lemica, gao2026meancache}, we introduce an exponent $\rho \ge 1.0$ to bias the NFE density toward the start of the reverse ODE:

\begin{equation}
    t_i = \mathcal{P}_{\text{sum}} \left( \mu_B \left( \left[ \frac{i}{B-1} \right]^\rho \right) \right).
\end{equation}

The projection operator $\mathcal{P}_{\text{sum}}$ rounds the samples to integers and redistributes residuals to the largest sampling gaps, ensuring the maximum timestep $T_{\text{max}}$ is strictly satisfied. This allows OTCache to faithfully ``morph'' high-fidelity structural priors into accelerated schedules for any target budget.

\section{Experiments}

\subsection{Experimental setup}

\subsubsection{Baselines and Compared Methods.}
To rigorously assess the effectiveness and generalizability of OTCache, we conduct experiments on three state-of-the-art generative frameworks: FLUX.1 [dev]~\cite{flux2024} for high-fidelity image synthesis, Qwen-Image~\cite{wu2025qwenimagetechnicalreport} for text-to-image generation, and HunyuanVideo~\cite{kong2024hunyuanvideo} for large-scale video generation. We benchmark OTCache against a comprehensive suite of representative caching paradigms. This includes threshold-based and structural redundancy methods, such as TaylorSeer~\cite{TaylorSeer2025}, DBCache~\cite{cache-dit@2025}, DiCache~\cite{bu2025dicache}, ToCa~\cite{zou2024accelerating}, and DuCa~\cite{zou2024DuCa}. We further compare with trajectory-aware acceleration techniques, including TeaCache~\cite{DBLP:journals/corr/abs-2411-19108}, as well as the latest Graph-based caching methods, LeMiCa~\cite{gao2025lemica} and MeanCache~\cite{gao2026meancache}.

\subsubsection{Metrics.} 
To rigorously assess the trade-off between computational acceleration and generation performance, we adopt a multi-faceted evaluation suite spanning efficiency, perceptual quality, and structural fidelity. Efficiency is quantified by total FLOPs and inference latency. For text-to-image (T2I) generation, we follow the DrawBench~\cite{saharia2022photorealistic} protocol and report ImageReward~\cite{xu2023imagereward} and CLIP Score~\cite{hessel2021clipscore} to evaluate high-level perceptual quality and semantic text-image alignment, respectively. For text-to-video (T2V) tasks, we employ VBench~\cite{huang2024vbench} to capture human-centric preferences across multiple temporal and spatial dimensions. Furthermore, to quantify the potential fidelity degradation introduced by the caching mechanism, we treat the full-step (uncompressed) inference as the reference ground truth and compute LPIPS~\cite{zhang2018unreasonable}, SSIM~\cite{wang2002universal}, and PSNR. These reconstruction metrics serve to measure the degree of perceptual similarity, structural consistency, and pixel-level precision maintained by OTCache relative to the original diffusion trajectory.

\subsubsection{Implementation Details.} 
All experiments are implemented in PyTorch and executed on NVIDIA H100 GPUs. To ensure a rigorous and fair evaluation, we adopt the sampling protocol established by prior works~\cite{DBLP:journals/corr/abs-2411-19108,gao2026meancache}, selecting 50 representative prompts (10 per attribute) from the T2V-CompBench~\cite{sun2024t2v} dataset. For all model architectures, we employ FlashAttention~\cite{dao2022flashattention} as the default attention backend to optimize memory bandwidth and computational throughput. In Stage 1 of OTCache, the reference budget is set to $B_{\mathrm{ref}}=20$. In Stage 2, the anchor search is conducted using Optuna with a budget of 200 trials to identify the optimal boundary caching sequence for 8-step inference. Notably, since TaylorSeer encounters out-of-memory (OOM) issues under HunyuanVideo, we uniformly adopt the cpu-offload setting to ensure fair comparison.

\subsection{Text-to-Image Generation}

As demonstrated in Table~\ref{tab:flux_compare} and Table~\ref{tab:qwen_compare}, \textbf{OTCache} consistently establishes a new state-of-the-art Pareto frontier on both FLUX.1 [dev] and Qwen-Image models. By introducing a caching mechanism from an Optimal Transport (OT) perspective, our method achieves a superior trade-off between acceleration ratio and generation fidelity compared to existing baselines.

\subsubsection{Performance on FLUX.1 [dev].} On the FLUX.1 [dev] model ($1024{\times}1024$), OTCache ($\mathcal{B}=15$) achieves a \textbf{$3.04\times$} speedup, surpassing MeanCache's $2.91\times$. Simultaneously, it further enhances reconstruction accuracy, reducing LPIPS from $0.142$ to \textbf{$0.126$} and boosting PSNR from $24.83$ to \textbf{$26.03$}. Notably, in high-acceleration regimes ($>3.6\times$), where baseline methods such as TeaCache experience catastrophic quality collapse, OTCache demonstrates remarkable robustness. At a significant \textbf{$4.50\times$} acceleration, it maintains an ImageReward of \textbf{$0.996$} and an LPIPS of \textbf{$0.254$}, outperforming MeanCache in both efficiency ($4.50\times$ vs. $4.12\times$) and perceptual quality.

\begin{table}[h]
\centering
\caption{Quantitative comparison of acceleration methods on \textbf{FLUX.1 [dev]} ($1024\times1024$). Best results are highlighted in \textbf{bold}.}
\label{tab:flux_compare}
\vspace{-2mm}
\setlength\tabcolsep{5pt}
\renewcommand{\arraystretch}{1.1}
\resizebox{\columnwidth}{!}{
\begin{tabular}{l | cc | ccccc}
\toprule
\multirow{2}{*}{\textbf{Method}} & \multicolumn{2}{c|}{\textbf{Acceleration}} & \multicolumn{5}{c}{\textbf{Visual Quality}} \\
\cmidrule(lr){2-3}\cmidrule(lr){4-8}
& \textbf{Latency(s) $\downarrow$} & \textbf{Speed $\uparrow$} & \textbf{Img Rew. $\uparrow$} & \textbf{CLIP $\uparrow$} & \textbf{LPIPS $\downarrow$} & \textbf{SSIM $\uparrow$} & \textbf{PSNR $\uparrow$} \\
\midrule
\textbf{Original: 50 steps} & 11.57 & 1.00$\times$ & 1.033 & 31.229 & -- & -- & -- \\
60\% steps & 7.01 & 1.65$\times$ & 0.984 & 31.242 & 0.217 & 0.808 & 20.26 \\
30\% steps & 3.60 & 3.21$\times$ & 0.880 & 30.832 & 0.399 & 0.682 & 15.80 \\
\midrule
TeaCache ($l=0.25$) & 4.62 & 2.50$\times$ & 0.960 & 31.145 & 0.338 & 0.721 & 17.29 \\
DiCache ($\delta=0.8$) & 4.32 & 2.68$\times$ & 0.675 & 30.814 & 0.416 & 0.717 & 21.27 \\
TaylorSeer ($\mathcal{N}=6,O=2$) & 4.24 & 2.74$\times$ & 0.971 & \textbf{31.310} & 0.415 & 0.663 & 16.28 \\
TaylorSeer ($\mathcal{N}=6,O=1$) & 4.06 & 2.85$\times$ & 0.961 & 31.191 & 0.419 & 0.660 & 15.83 \\
LeMiCa ($\mathcal{B}=15$) & 4.13 & 2.80$\times$ & 0.991 & 31.125 & 0.153 & 0.858 & 24.45 \\
MeanCache ($\mathcal{B}=15$) & 3.98 & 2.91$\times$ & 1.010 & 31.244 & 0.142 & 0.870 & 24.83 \\
\rowcolor{blue!15}
\textbf{OTCache ($\mathcal{B}=15$)} & \textbf{3.80} & \textbf{3.04}$\times$ & \textbf{1.011} & 31.167 & \textbf{0.126} & \textbf{0.881} & \textbf{26.03} \\
\midrule
TeaCache ($l=1.5$) & 3.16 & 3.66$\times$ & 0.717 & 30.696 & 0.504 & 0.624 & 15.01 \\
% DiCache ($\delta=2.0$) & 3.14 & 3.68$\times$ & -0.652 & 27.613 & 0.586 & 0.588 & 17.45 \\
% TaylorSeer ($\mathcal{N}=20,O=1$) & 3.10 & 3.73$\times$ & -0.727 & 24.412 & 0.798 & 0.443 & 11.22 \\
LeMiCa ($\mathcal{B}=10$) & 3.21 & 3.60$\times$ & 0.981 & \textbf{31.355} & 0.312 & 0.740 & 19.03 \\
MeanCache ($\mathcal{B}=10$) & 2.81 & 4.12$\times$ & 0.993 & 31.323 & 0.272 & 0.761 & 19.43 \\
\rowcolor{blue!15}
\textbf{OTCache ($\mathcal{B}=10$)} & \textbf{2.57} & \textbf{4.50$\times$} & \textbf{0.996} & 31.269 & \textbf{0.254} & \textbf{0.780} & \textbf{20.22} \\
\bottomrule
\end{tabular}}
\vspace{-5mm}
\end{table}

\begin{figure*}[h]
  \centering
  \vspace{-5mm}
  \includegraphics[width=0.9\linewidth]{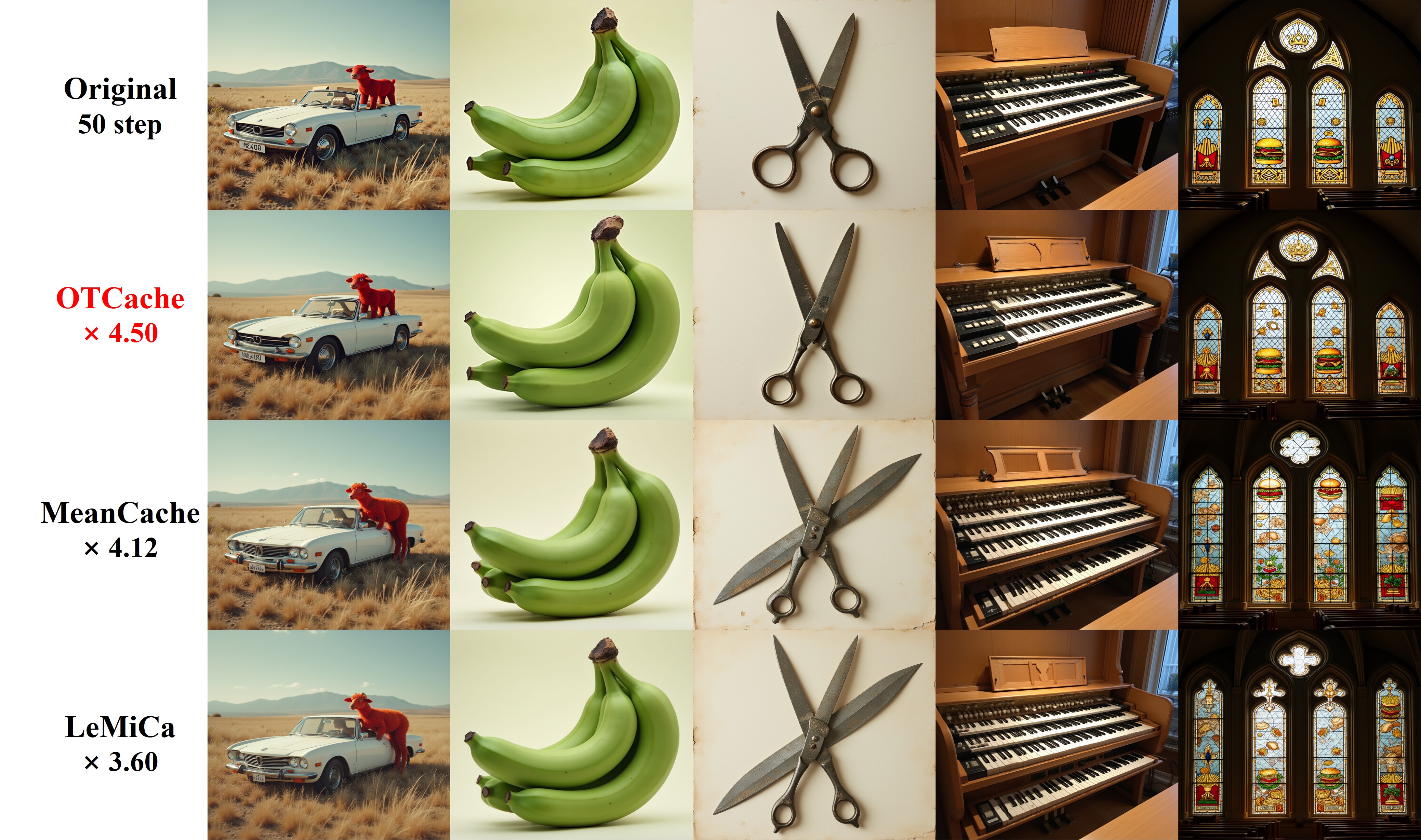}
  \vspace{-2pt}
  \caption{Comparison of different methods at high acceleration ratios on \textbf{FLUX.1 [dev]} (1024×1024).}
  \vspace{-10mm}
  \label{fig:flux_compare}
\end{figure*}

As shown in Fig.~\ref{fig:flux_compare}, on FLUX.1 [dev], when the acceleration exceeds $3.6\times$, baseline methods exhibit more frequent content inconsistencies and object distortions (e.g., malformed legs of the red goat, unnatural scissors, and redundant piano keys). In contrast, OTCache achieves better content consistency and higher visual quality at an even larger speedup ($4.50\times$), substantially outperforming graph-based caching baselines.

\begin{table}[h]
\centering
\caption{Quantitative comparison of acceleration methods on \textbf{Qwen-Image} ($1664\times928$). Best results are highlighted in \textbf{bold}.}
\label{tab:qwen_compare}
\setlength\tabcolsep{5pt}
\renewcommand{\arraystretch}{1.1}
\resizebox{\columnwidth}{!}{
\begin{tabular}{l | cc | ccccc}
\toprule
\multirow{2}{*}{\textbf{Method}} & \multicolumn{2}{c|}{\textbf{Acceleration}} & \multicolumn{5}{c}{\textbf{Visual Quality}} \\
\cmidrule(lr){2-3}\cmidrule(lr){4-8}
& \textbf{Latency(s) $\downarrow$} & \textbf{Speed $\uparrow$} & \textbf{Img Rew. $\uparrow$} & \textbf{CLIP $\uparrow$} & \textbf{LPIPS $\downarrow$} & \textbf{SSIM $\uparrow$} & \textbf{PSNR $\uparrow$} \\
\midrule
\textbf{Original: 50 steps} & 32.68 & 1.00$\times$ & 1.180 & 33.626 & -- & -- & -- \\
30\% steps & 9.86 & 3.31$\times$ & 1.128 & 33.026 & 0.363 & 0.727 & 15.83 \\
\midrule
TeaCache ($l=0.6$) & 18.52 & 1.76$\times$ & 1.087 & 32.598 & 0.416 & 0.698 & 14.90 \\
DBCache ($r=0.6$) & 11.92 & 2.74$\times$ & 1.016 & 33.435 & 0.298 & 0.825 & 22.22 \\
LeMiCa ($\mathcal{B}=15$) & 13.26 & 2.46$\times$ & 1.120 & 33.590 & 0.122 & 0.924 & 26.89 \\
MeanCache ($\mathcal{B}=15$) & 11.45 & 2.85$\times$ & 1.159 & 33.636 & 0.075 & 0.938 & 27.66 \\
\rowcolor{blue!15}
\textbf{OTCache ($\mathcal{B}=15$)} & \textbf{10.21} & \textbf{3.20}$\times$ & \textbf{1.164} & \textbf{33.652} & \textbf{0.069} & \textbf{0.943} & \textbf{28.12} \\
\midrule
LeMiCa ($\mathcal{B}=13$) & 11.54 & 2.83$\times$ & 1.096 & 33.667 & 0.177 & 0.884 & 24.06 \\
MeanCache ($\mathcal{B}=13$) & 9.09 & 3.60$\times$ & 1.147 & \textbf{33.799} & 0.113 & 0.907 & 24.80 \\
\rowcolor{blue!15}
\textbf{OTCache ($\mathcal{B}=13$)} & \textbf{8.87} & \textbf{3.68}$\times$ & \textbf{1.150} & 33.614 & \textbf{0.087} & \textbf{0.930} & \textbf{26.79} \\
\midrule
LeMiCa ($\mathcal{B}=10$) & 10.78 & 3.03$\times$ & 1.111 & \textbf{33.739} & 0.253 & 0.816 & 19.09 \\
MeanCache ($\mathcal{B}=10$) & 7.16 & 4.56$\times$ & 1.142 & 33.621 & 0.236 & 0.815 & 18.98 \\
\rowcolor{blue!15}
\textbf{OTCache ($\mathcal{B}=10$)} & \textbf{6.95} & \textbf{4.70}$\times$ & \textbf{1.147} & 33.584 & \textbf{0.171} & \textbf{0.864} & \textbf{21.48} \\
\bottomrule
\end{tabular}}
\vspace{-7mm}
\end{table}

\subsubsection{Performance on Qwen-Image.} For high-resolution generation on Qwen-Image ($1664{\times}928$, 16:9 aspect ratio), the efficiency gains of OTCache are even more pronounced. At $\mathcal{B}=15$, it reaches a \textbf{$3.20\times$} speedup with a near-lossless LPIPS of \textbf{$0.069$}, which is significantly superior to MeanCache's $0.075$ at $2.85\times$. Even under an extreme \textbf{$4.70\times$} acceleration ($\mathcal{B}=10$), OTCache preserves high structural fidelity with an SSIM of \textbf{$0.864$} and a PSNR of \textbf{$21.48$}, whereas other caching-based baselines either provide lower speedups or incur significantly higher reconstruction errors.

\begin{figure*}[h]
  \centering
  \vspace{-5mm}
  \includegraphics[width=0.85\linewidth]{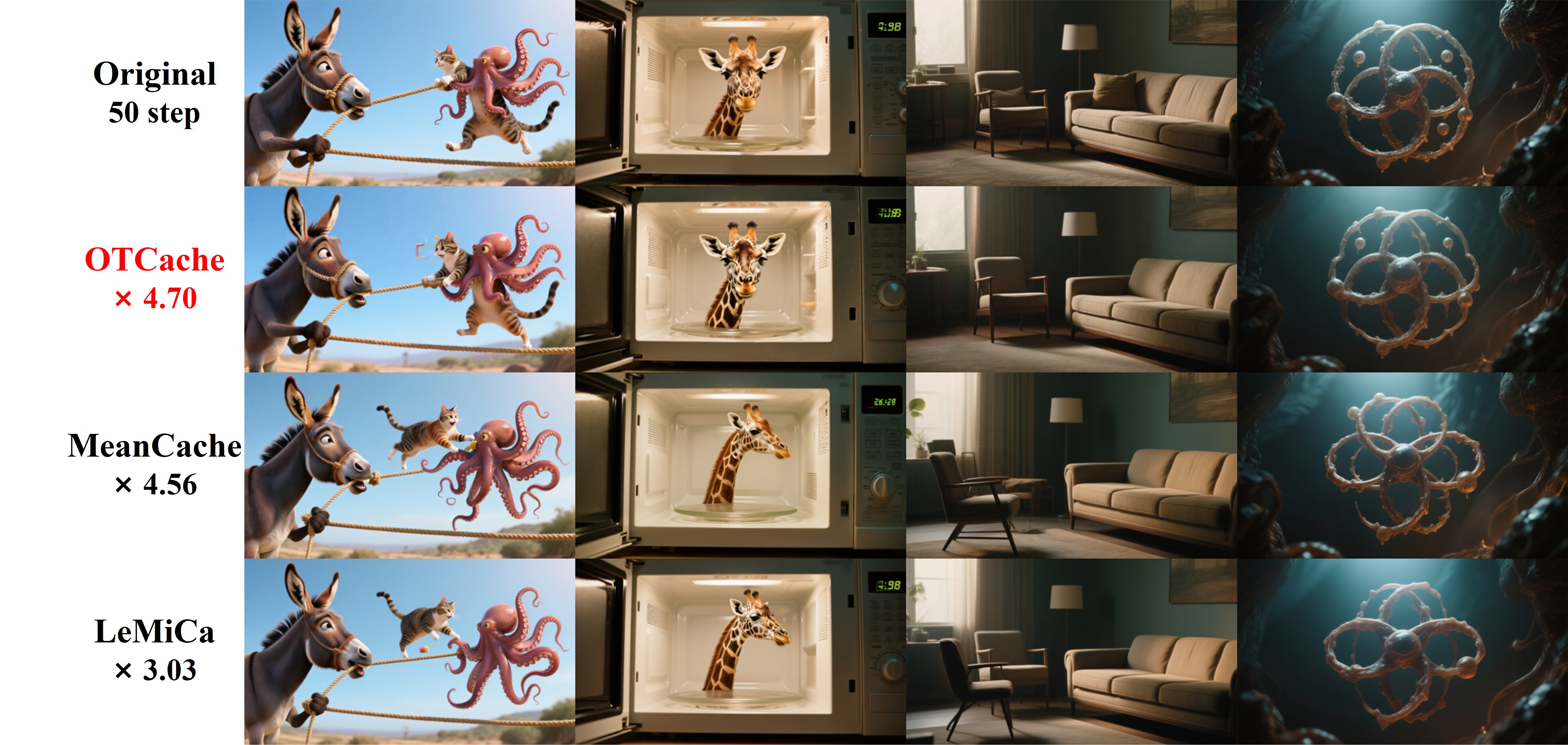}
  \vspace{-2pt}
  \caption{Comparison of different methods at high acceleration ratios on \textbf{Qwen-Image} (1664×928).}
  \vspace{-5mm}
  \label{fig:qw_compare}
\end{figure*}

Similar to the observations on FLUX.1 [dev], the qualitative comparisons in Fig.~\ref{fig:qw_compare} further highlight the strong generalization capability of OTCache for text-to-image models. In the 16:9 generation setting commonly used for poster-style compositions, OTCache maintains significantly better content consistency before and after acceleration. In contrast, graph-based caching baselines such as MeanCache and LeMiCa exhibit noticeable content shifts, including changes in the relative positions of objects (e.g., the cat and the octopus) and altered orientations of subjects (e.g., the giraffe). These qualitative findings are consistent with the analysis in Sec.~\ref{sec:rethink} and the quantitative results reported in Table~\ref{tab:qwen_compare}.

\subsection{Text-to-Video Generation.}

As shown in Table~\ref{tab:hunyuan_compare}, OTCache establishes a new state-of-the-art for efficient text-to-video generation on HunyuanVideo. By modeling cache scheduling as a budget-conditioned path evolution via Optimal Transport (OT), OTCache achieves a $3.21\times$ speedup, matching TeaCache and surpassing MeanCache’s $3.05\times$, while improving reconstruction fidelity (LPIPS $0.176\!\rightarrow\!0.162$).

\begin{figure*}[h]
  \centering
  \vspace{-2pt}
  \includegraphics[width=0.8\linewidth]{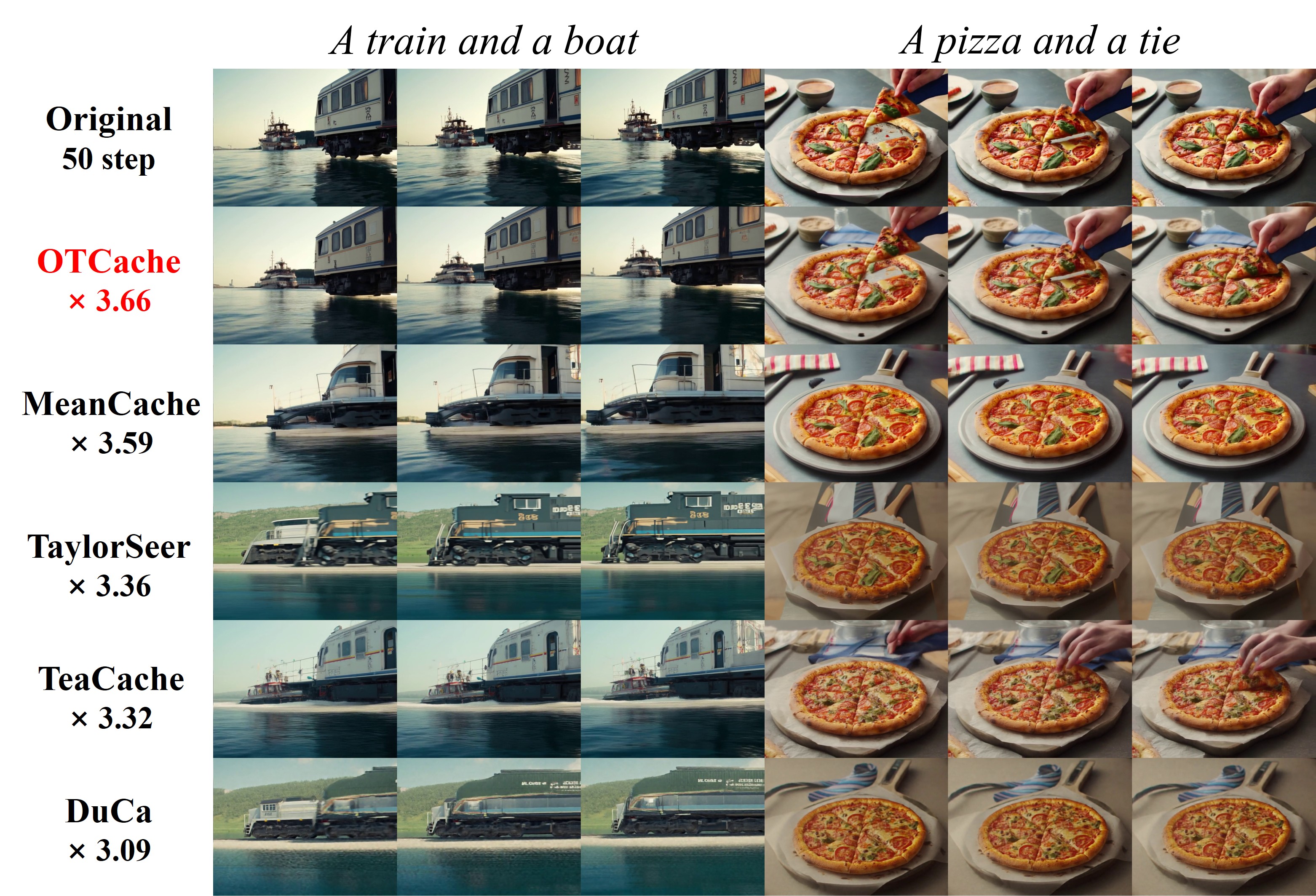}
  \vspace{-2pt}
  \caption{Comparison of different methods at high acceleration ratios on \textbf{HunyuanVideo}.}
  \vspace{-5mm}
  \label{fig:compare_video}
\end{figure*}

OTCache also remains robust under extreme acceleration. At $3.66\times$, it achieves a VBench score of $80.37\%$ with an LPIPS of $0.252$, consistently outperforming MeanCache in both efficiency and visual quality, while prior methods such as DiCache and TeaCache exhibit noticeable temporal artifacts. As illustrated in Fig.~\ref{fig:compare_video}, we further present qualitative comparisons on representative key frames. OTCache consistently produces more stable structures and clearer visual details, whereas competing methods show more distortions and temporal inconsistencies. Fig.~\ref{fig:radar_vbench} further provides a multi-dimensional comparison across VBench metrics.
\begin{table*}[h]
\centering
\vspace{-15pt}
\caption{Quantitative comparison in text-to-video generation on \textbf{HunyuanVideo}. Best results are in \textbf{bold}.}
\vspace{-10pt}
\label{tab:hunyuan_compare}
\setlength\tabcolsep{6pt} % 略微增加间距提升可读性
\renewcommand{\arraystretch}{1.1}
\resizebox{0.95\textwidth}{!}{
\begin{tabular}{l | cc | cccc}
\toprule
\multirow{2}{*}{\textbf{Method}} & \multicolumn{2}{c|}{\textbf{Acceleration}} & \multicolumn{4}{c}{\textbf{Visual Quality}} \\
\cmidrule(lr){2-3}\cmidrule(lr){4-7}
& \textbf{Latency(s) $\downarrow$} & \textbf{Speed $\uparrow$} & \textbf{VBench $\uparrow$} & \textbf{LPIPS $\downarrow$} & \textbf{SSIM $\uparrow$} & \textbf{PSNR $\uparrow$} \\
\midrule
\textbf{Original: 50 steps} & 105.92 & 1.00$\times$ & 80.39\% & -- & -- & -- \\
30\% steps & 39.53 & 2.68$\times$ & 79.84\% & 0.381 & 0.659 & 17.335 \\
\midrule
ToCa ($\mathcal{N}=5$) & 36.17 & 2.93$\times$ & 79.51\% & 0.454 & 0.590 & 15.765 \\
Duca ($\mathcal{N}=5$) & 34.32 & 3.09$\times$ & 79.54\% & 0.454 & 0.595 & 15.807 \\
DiCache ($\delta=0.8$) & 33.76 & 3.11$\times$ & 74.09\% & 0.382 & 0.701 & 22.053 \\
TaylorSeer ($\mathcal{N}=5,O=1$) & 34.95 & 3.03$\times$ & 79.95\% & 0.428 & 0.603 & 16.026 \\
Teacache ($l=0.33$) & 34.06 & 3.11$\times$ & 80.02\% & 0.363 & 0.651 & 17.957 \\
MeanCache ($\mathcal{B}=12$) & 33.05 & 3.21$\times$ & 80.01\% & 0.176 & 0.809 & 24.002 \\
\rowcolor{blue!15}
\textbf{OTCache ($\mathcal{B}=12$)} & \textbf{33.01} & \textbf{3.21$\times$} & \textbf{80.20\%} &\textbf{0.162} & \textbf{0.815} & \textbf{24.356} \\
\midrule
DiCache ($\delta=3.0$) & 31.81 & 3.33$\times$ & 70.86\% & 0.583 & 0.490 & 19.098 \\
Teacache ($l=0.39$) & 31.86 & 3.32$\times$ & 79.75\% & 0.396 & 0.631 & 17.382 \\
TaylorSeer ($\mathcal{N}=7,O=1$) & 31.50 & 3.36$\times$ & 79.76\% & 0.480 & 0.595 & 15.444 \\

MeanCache ($\mathcal{B}=10$) & 29.48 & 3.59$\times$ & 80.08\% & 0.269 & 0.732 & 20.464 \\
\rowcolor{blue!15}
\textbf{OTCache ($\mathcal{B}=10$)} & \textbf{28.96} & \textbf{3.66$\times$} & \textbf{80.37\%} & \textbf{0.252} & \textbf{0.746} & \textbf{21.150} \\
\bottomrule
\end{tabular}}
\vspace{-10mm}
\end{table*}

\subsection{Ablation Study}

\subsubsection{Content Consistency}
\begin{figure*}[h]
  \centering
  \begin{minipage}[b]{0.35\linewidth}
    \centering
    \includegraphics[width=\linewidth]{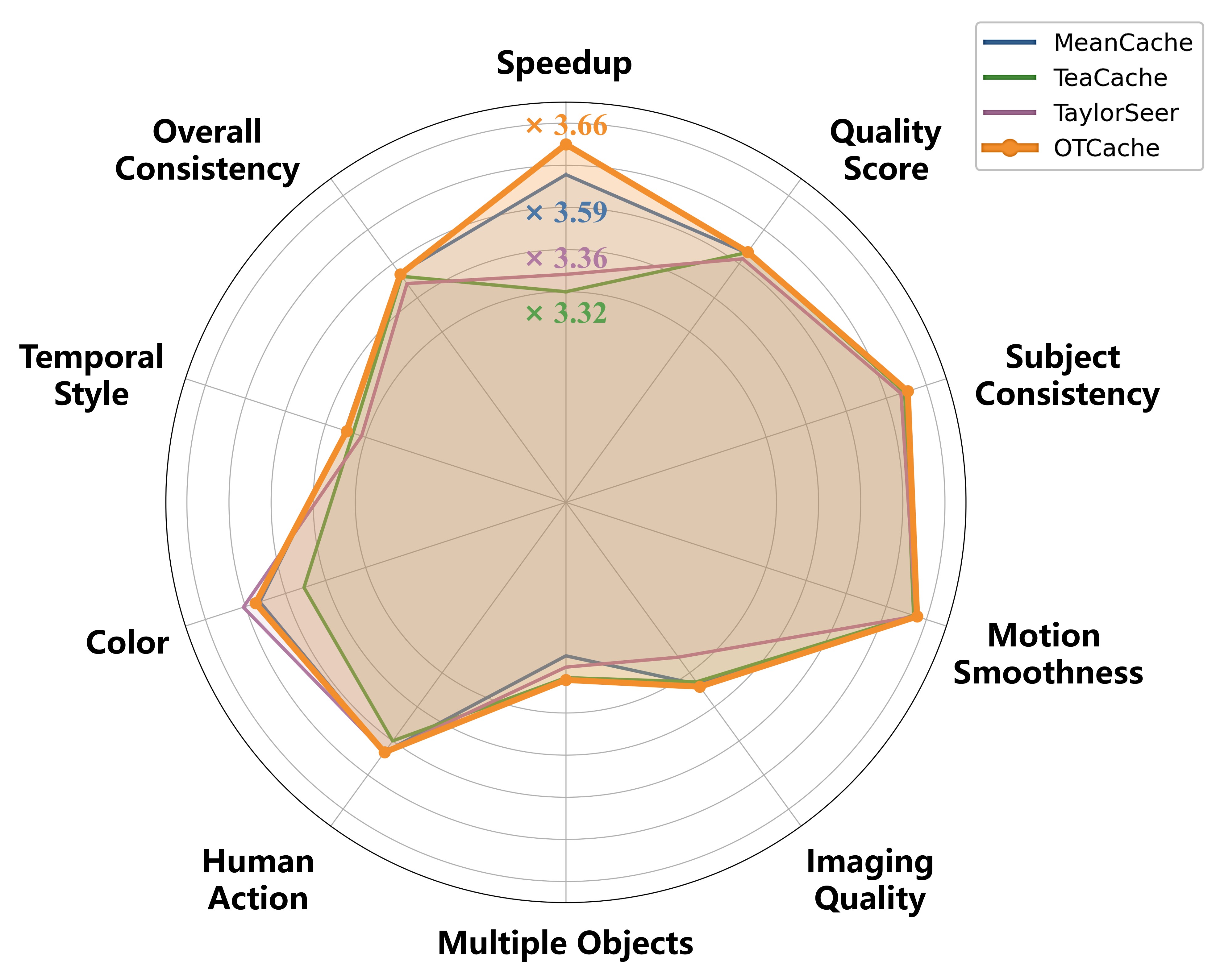}
    \vspace{-15pt} % 缩小图片与标题间距
    \caption{VBench metrics and acceleration ratio of proposed OTCache and other methods.}
    \label{fig:radar_vbench}
  \end{minipage}
  \hfill % 撑开中间空白
  \begin{minipage}[b]{0.62\linewidth}
    \centering
    \includegraphics[width=\linewidth]{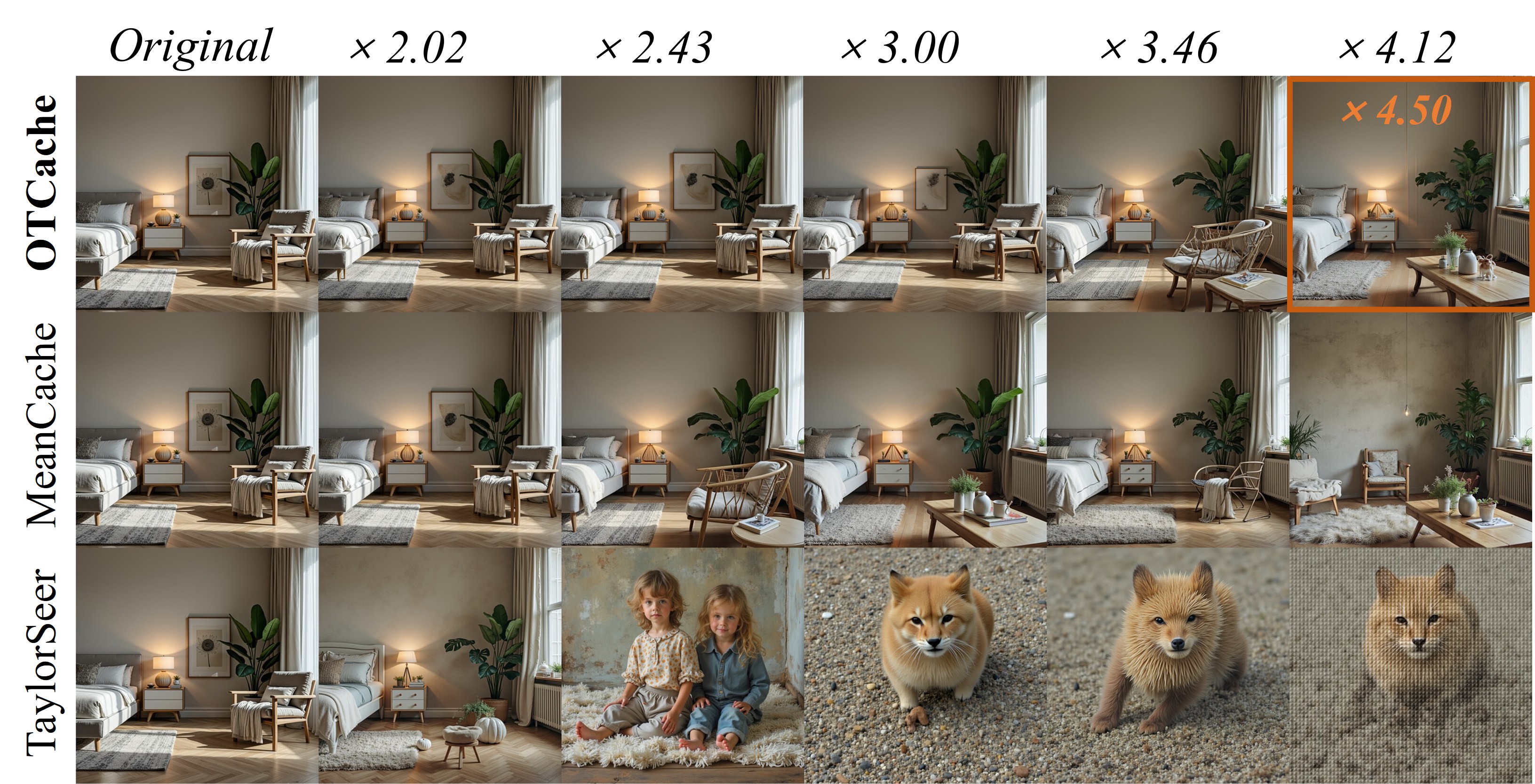}
    \vspace{-15pt}
    \caption{Content consistency under a rare-word prompt (“Matutinal”) across varying acceleration ratios}
    \label{fig:content_consistency}
  \end{minipage}
  \vspace{-4mm}
\end{figure*}

Following MeanCache, we conduct a detailed evaluation of OTCache on rare-word generation, where semantic ambiguity and low-frequency usage pose significant challenges to text-to-image models. As shown in Figure~\ref{fig:content_consistency}, MeanCache outperforms TeaCache in maintaining content consistency under moderate acceleration. However, as the acceleration ratio increases, MeanCache gradually exhibits content drift: an extra round table appears at $2.43\times$, chairs are missing at $3.00\times$, and the bedside lamp disappears at $4.12\times$, indicating degradation of fine-grained semantic fidelity. In contrast, OTCache preserves content more faithfully across all acceleration levels. Notably, even under a more extreme $4.50\times$ speedup (orange box), OTCache still retains critical elements such as the bedside lamp and its illumination, demonstrating superior robustness in maintaining rare-word semantics and structural details under aggressive acceleration.

\subsubsection{Effect of $\rho$}
The parameter $\rho$ modulates the NFE density, where higher values prioritize the early-stage reverse ODE to stabilize high-variance sampling. We evaluate $\rho \in [1.0, 1.45]$ at a fixed budget $\mathcal{B}=15$ (Table~\ref{tab:rho_ablation}). Compared to uniform sampling ($\rho=1.0$), non-linear allocation significantly enhances reconstruction fidelity. Specifically, $\rho=1.3$ achieves the optimal balance, yielding the highest PSNR ($26.034$) and lowest LPIPS ($0.126$). While $\rho=1.0$ marginally leads in Image Reward, its inferior PSNR and LPIPS indicate structural instability. Conversely, $\rho=1.45$ causes performance degradation, suggesting that excessive early-stage bias compromises late-stage refinement. Thus, we fix \textbf{$\rho=1.3$} for all subsequent experiments.

\begin{table*}[t]
  \centering
  % --- 左表 ---
  \begin{minipage}[t]{0.48\linewidth}
    \centering
    \caption{Search accuracy on LPIPS $\downarrow$.}
     \vspace{-5.5 pt}
    \label{tab:search_gain}
    \small
    \addtolength{\tabcolsep}{3pt} % 压缩列间距以放下 4 列数据
    \begin{tabular}{c c c c}
    \toprule
    Rank & MC & Search & Gain (\%) \\
    \midrule
    Top-1 & 0.338 & \textbf{0.261} & \textbf{25.04\%} \\
    Top-2 & 0.338 & 0.263 & 24.48\% \\
    Top-3 & 0.338 & 0.263 & 24.26\% \\
    Top-4 & 0.338 & 0.264 & 24.02\% \\
    Top-5 & 0.338 & 0.265 & 23.67\% \\
    \bottomrule
    \end{tabular}
  \end{minipage}
  \hfill
  % --- 右表 ---
  \begin{minipage}[t]{0.48\linewidth}
    \centering
    \caption{Impact of parameter $\rho$ on quality metrics.}
     \vspace{-6 pt}
    \label{tab:rho_ablation}
    \small
    \begin{tabular}{c c c c c}
    \toprule
    $\rho$ & Img Rwd. $\uparrow$ & PSNR $\uparrow$ & SSIM $\uparrow$ & LPIPS $\downarrow$ \\
    \midrule
    1.00 & \textbf{1.022} & 24.304 & 0.864 & 0.139 \\
    1.15 & 1.000          & 24.847 & 0.871 & 0.134 \\
    1.30 & 1.011          & \textbf{26.034} & \textbf{0.880} & \textbf{0.126} \\
    1.45 & 1.001          & 25.798 & 0.877 & 0.131 \\
    \bottomrule
    \end{tabular}
  \end{minipage}
  \vspace{-2mm}
\end{table*}

\subsubsection{Effectiveness and Efficiency of Anchor Search}

\begin{figure}[t]
  \centering
  \includegraphics[width=0.75\linewidth]{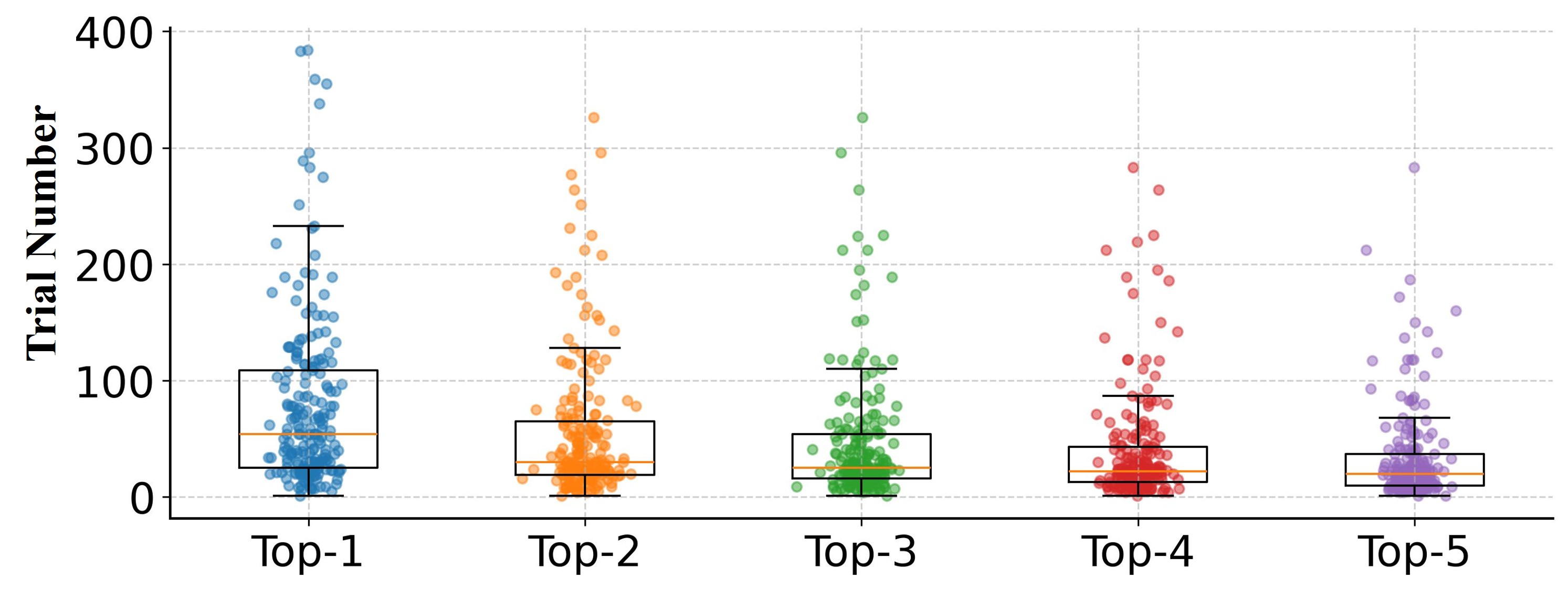}
    \caption{\textbf{Search efficiency}. Number of trials required to identify Top-$K$ schedules across 50 prompt-seed pairs at budget $B=8$. The median convergence for the Top-1 optimum is around 50 trials, with most near-optimal schedules discovered within 100 trials.}
  \label{fig:optuna_topk}
  \vspace{-7mm}
\end{figure}

We analyze the effectiveness and efficiency of the anchor search under $B_{\mathrm{anc}}=8$ across 100 prompt-seed pairs. Table~\ref{tab:search_gain} shows that the searched anchors consistently outperform the MeanCache (MC) baseline. The Top-1 schedule reduces LPIPS by \textbf{25.04\%} on average, while even the Top-5 candidates maintain more than \textbf{23\%} improvement over MC, indicating that directly optimizing the end-to-end perceptual objective effectively recovers the fidelity loss introduced by surrogate-based graph methods. Meanwhile, Fig.~\ref{fig:optuna_topk} demonstrates that the search remains highly efficient. The median number of trials required to identify the Top-1 optimum is around 50, and the vast majority of cases converge within 200 trials. These results show that the anchor search in Stage 2 of OTCache is both efficient and effective, providing a reliable boundary condition for schedule prediction with minimal computational overhead.

\section{Conclusion}

This paper introduces \textbf{OTCache}, a training-free framework for accelerating diffusion sampling through caching schedule prediction. Unlike existing graph-based caching methods that rely on additive shortest-path objectives, OTCache models the evolution of caching schedules across inference budgets as a smooth trajectory in policy space inspired by Optimal Transport. The framework consists of three stages: a high-fidelity reference schedule under a conservative budget, an anchor schedule obtained via lightweight end-to-end search in the ultra-low NFE regime, and a quantile-interpolation strategy that predicts schedules for arbitrary budgets through continuous warping representations. More broadly, OTCache provides a new perspective on diffusion acceleration through geometry-aware schedule modeling. 

\section*{Acknowledgements}
This work was supported by the National Natural Science Foundation of China Enterprise Innovation and Development Joint Fund Project under Grant U24B20181. 

% \clearpage\mbox{}Page \thepage\ of the manuscript.
% \clearpage\mbox{}Page \thepage\ of the manuscript.
% \clearpage\mbox{}Page \thepage\ of the manuscript.
% \clearpage\mbox{}Page \thepage\ of the manuscript.
% \clearpage\mbox{}Page \thepage\ of the manuscript. This is the last page.
% Now we have reached the maximum length of an ECCV \ECCVyear{} submission (excluding references and acknowledgements).
% References should start immediately after the main text, but can continue past p.\ 14 if needed. 
\clearpage 
 % TODO FINAL: This \clearpage needs to be removed from both review and camera-ready versions.

% \section*{Acknowledgements}
% Please insert your acknowledgments here.


\bibliographystyle{splncs04}
\bibliography{main}

\clearpage  

\appendix

\centering \textbf{\Large OTCache: Optimal Transport for Geometry-Aware Caching in Diffusion Models}

\vspace{4mm}
\centering {\Large Appendix}

\raggedright

\justifying %

\vspace{7mm}

\section{Baselines and Experimental Settings}
\label{sec:baselines_settings}

We evaluate OTCache across text-to-image and text-to-video tasks using three representative models. Detailed baseline configurations are as follows:

\begin{itemize}
    \item FLUX.1 [dev]: Two distinct acceleration regimes are considered to assess performance stability. 
    (1) For \textit{moderate acceleration} ($2.50\times$--$3.04\times$), the accumulation error threshold is set to $l=0.25$ for TeaCache \cite{DBLP:journals/corr/abs-2411-19108} and the control factor $\delta=0.8$ for DiCache \cite{bu2025dicache}. For TaylorSeer \cite{TaylorSeer2025}, which reformulates cache reuse as cache prediction, the settings $\mathcal{N}=6, O=1$ and $\mathcal{N}=6, O=2$ are adopted. Graph-based caching methods are configured with a fixed computation budget of $\mathcal{B}=15$. 
    (2) For \textit{high acceleration} ($>3.6\times$), where traditional threshold-based methods often suffer from catastrophic quality degradation, the evaluation focuses on comparing generative quality and perceptual metrics under a constrained budget of $\mathcal{B}=10$.

    \item Qwen-Image: DBCache \cite{cache-dit@2025} is first introduced, which adjusts acceleration via the ratio $r$, using default hyperparameters $F_n=2$ and $B_n=4$. To provide a granular analysis of graph-based caching frameworks, detailed comparisons are conducted across varying computation budgets, specifically $\mathcal{B} \in \{15, 13, 10\}$. This facilitates a rigorous evaluation of the Pareto frontier of the proposed OTCache against existing graph-optimal scheduling strategies.

    \item HunyuanVideo: In addition to the aforementioned baselines, comparisons are conducted with ToCa \cite{zou2024accelerating} and DuCa \cite{zou2024DuCa}, both configured with a caching interval of $\mathcal{N}=5$ to align with the TaylorSeer setting. Given the significant computational demands of video generation, the performance of graph-based methods under budgets of $\mathcal{B}=12$ and $\mathcal{B}=10$ is particularly reported. 
\end{itemize}

\section{Search Space Analysis}
\label{sec:search_space}

\begin{figure}[h]
    \centering
    \includegraphics[width=0.65\linewidth]{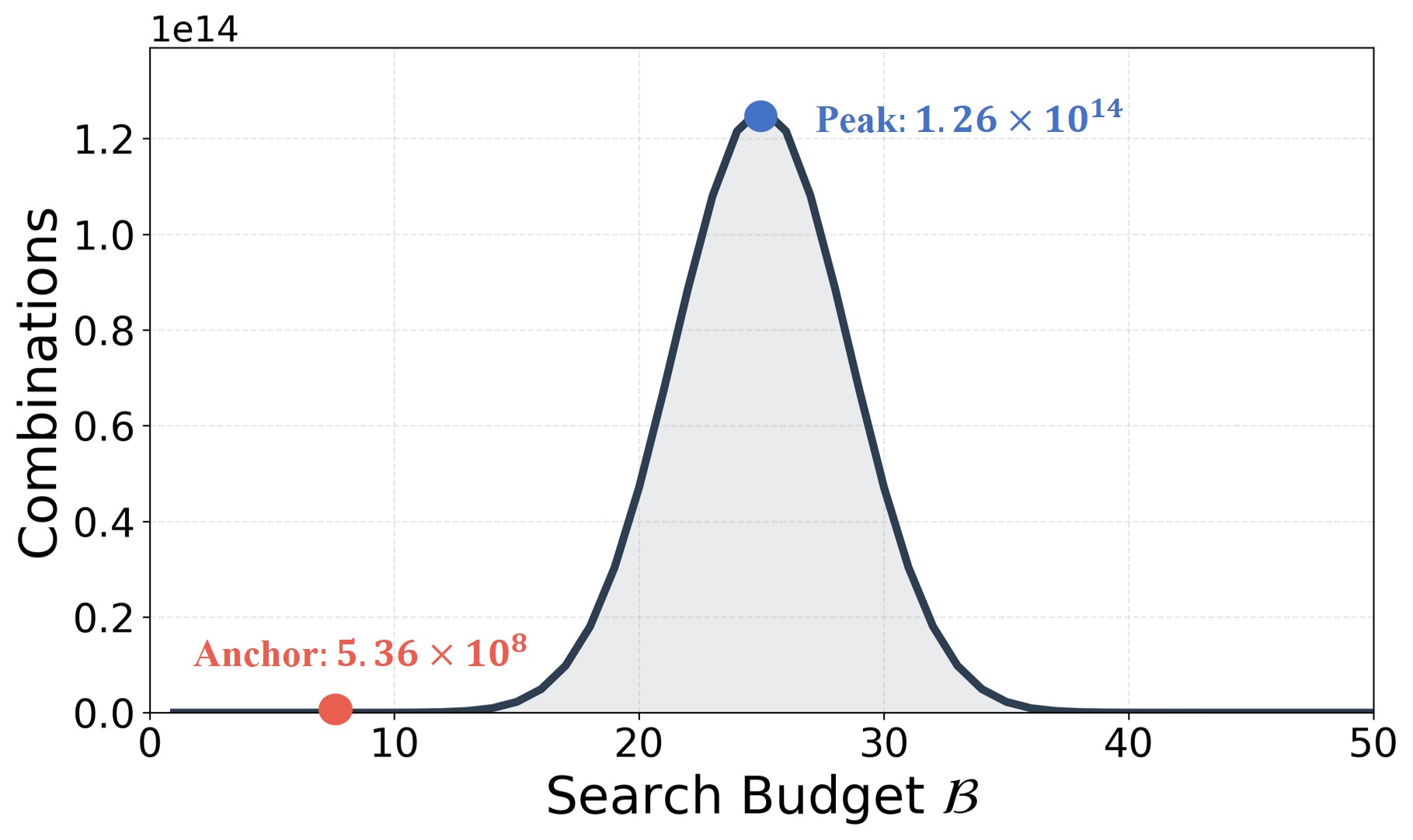}
    \caption{Combinatorial search space analysis for $T=50$. The search complexity peaks at $\mathcal{B}=25$, while our chosen anchor at $\mathcal{B}=8$ resides in a significantly more tractable region, enabling efficient policy optimization.}
    \label{fig:search_space}
\end{figure}

The optimization of a caching schedule is inherently a combinatorial problem, where the search space for selecting $\mathcal{B}$ steps from $T$ timesteps is bounded by $\binom{T}{\mathcal{B}}$. As illustrated in Fig.~\ref{fig:search_space}, this complexity exhibits a massive explosion as $\mathcal{B}$ approaches $T/2$. For a standard $T=50$ configuration, searching at a mid-range budget (e.g., $\mathcal{B}=25$) involves an immense space of $\approx 1.26 \times 10^{14}$ candidates, rendering global optimization computationally prohibitive. In contrast, OTCache strategically performs its lightweight search at a low-budget anchor point (e.g., $\mathcal{B}=8$), where the search space ($\approx 5.36 \times 10^{8}$) is significantly more tractable for the Optuna-based optimizer. By establishing this robust anchor policy $\pi_{anc}$, schedules for any intermediate budgets can be accurately predicted via Optimal Transport interpolation, effectively bypassing the "curse of dimensionality" at the peak search complexity regimes.

% \vspace{-5mm}

% \vspace{-5mm}

\section{Effect of Anchor Budget}
\label{sec:anchor_selection}

The anchor budget $\mathcal{B}_{anc}$ governs the trade-off between search tractability and predictive stability. Although a smaller $\mathcal{B}_{anc}$ constrains the search space complexity, it risks introducing significant approximation errors during policy interpolation. We investigate this by predicting the $\mathcal{B}=10$ schedule using anchor lengths $\mathcal{B}_{anc} \in \{8, 6, 4\}$.As summarized in Tab.~\ref{tab:anchor_ablation}, visual quality exhibits a clear degradation as $\mathcal{B}_{anc}$ decreases. Specifically, $\mathcal{B}_{anc}=8$ achieves the highest fidelity, with a PSNR of 20.22 and an LPIPS of 0.254. At lower budgets (i.e., 6 or 4), sampling trajectories become inherently unstable, failing to provide sufficient geometric priors for accurate quantile-based interpolation. Consequently, we adopt $\mathcal{B}_{anc}=8$ for standard 50-step sampling to maintain a robust balance between optimization efficiency and reconstruction quality.

\begin{table}[h]
\centering
\caption{Ablation of anchor budget $\mathcal{B}_{anc}$ for predicting $\mathcal{B}=10$ schedules.}
\label{tab:anchor_ablation}
\vspace{-2mm}
\setlength\tabcolsep{8pt}
\renewcommand{\arraystretch}{1.1}
\resizebox{\columnwidth}{!}{
\begin{tabular}{l | ccccc}
\toprule
  & \textbf{Img Rew. $\uparrow$} & \textbf{CLIP $\uparrow$} & \textbf{LPIPS $\downarrow$} & \textbf{SSIM $\uparrow$} & \textbf{PSNR $\uparrow$} \\
\midrule
% --- 数据支持  ---
$\mathcal{B}_{anc}=8$ & \textbf{0.996} & 31.269 & \textbf{0.254} & \textbf{0.780} & \textbf{20.22} \\
$\mathcal{B}_{anc}=6$           & 0.978          & \textbf{31.326} & 0.301          & 0.752          & 18.59          \\
$\mathcal{B}_{anc}=4$           & 0.970          & 31.154          & 0.302          & 0.753          & 18.29          \\
\bottomrule
\end{tabular}}
\vspace{-4mm}
\end{table}

\section{Impact of Initialization}
\label{sec:init_strategies}

\begin{figure}[h]
    \centering
    \includegraphics[width=0.8\linewidth]{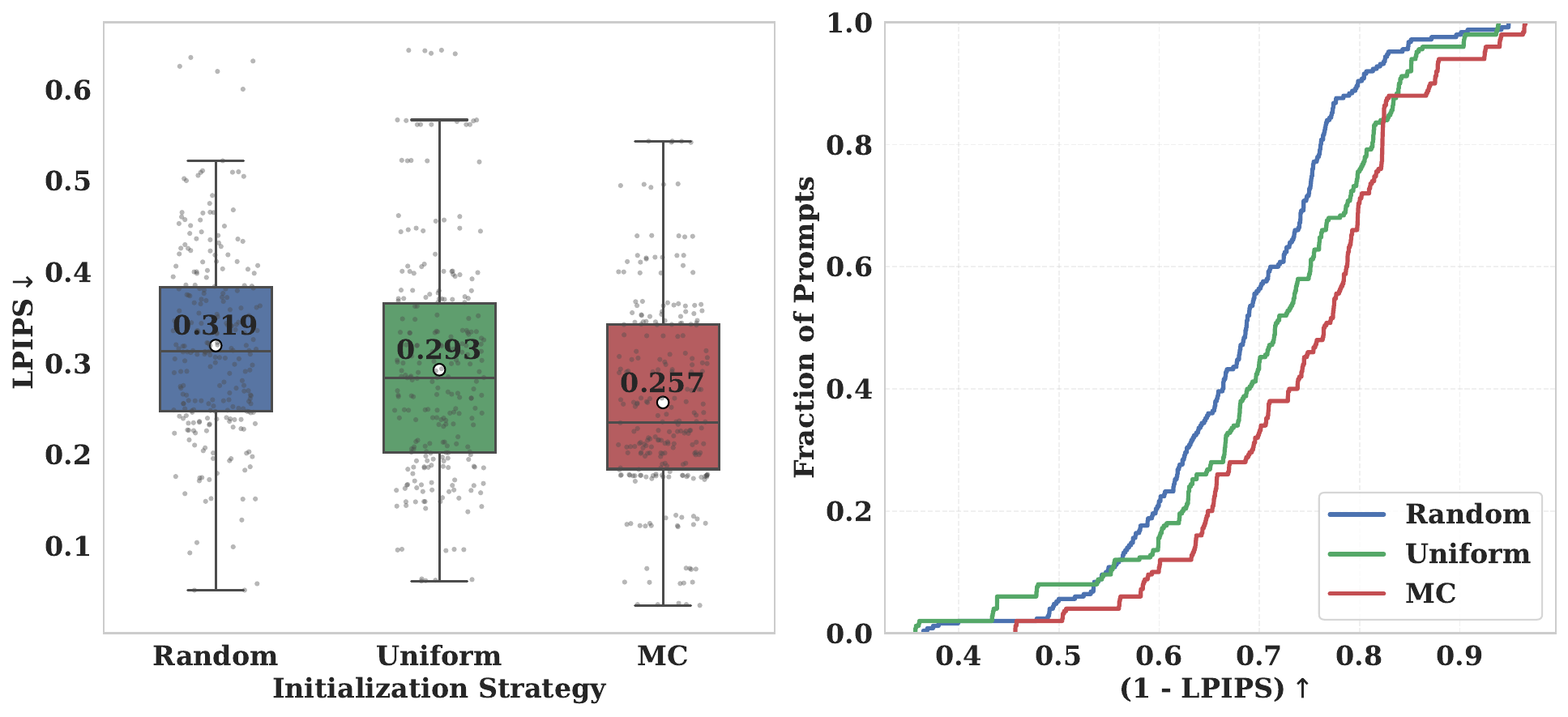}
    \caption{Evaluation of search initialization strategies at $\mathcal{B}=8$. MC achieves the lowest average LPIPS among 250 samples (left), with its ECDF curve (right) showing a distinct rightward shift, indicating superior robustness in discovering high-fidelity caching policies.}
    \label{fig:search_strategies}
\end{figure}

Existing acceleration methods in low-NFE regimes provide valuable structural priors that can enhance the optimization process. To demonstrate this, we investigate whether utilizing these baseline strategies as search initializations leads to superior policy discovery at a computation budget of $\mathcal{B}=8$. We compare three initialization approaches: \textit{Random}, \textit{Uniform}, and \textit{MeanCache (MC)}. For each strategy, the top-5 caching schedules (ranked by LPIPS) are extracted across 50 diverse prompts, resulting in 250 experimental samples for robust statistical analysis. As illustrated in Fig.~\ref{fig:search_strategies}, the initialization strategy significantly dictates the final optimization performance. The box plot (left) reveals that MC-based initialization consistently outperforms other baselines, achieving a superior average LPIPS of 0.257, compared to 0.319 for Random and 0.293 for Uniform. This performance gap is further elucidated by the Empirical Cumulative Distribution Function (ECDF) on the right, where the MC curve (red) exhibits a distinct rightward shift. This trend demonstrates that a substantially larger fraction of samples achieves high reconstruction fidelity ($1-\text{LPIPS} \uparrow$) under MC guidance. Consequently, OTCache adopts MC ($\mathcal{B}=8$) as the default initialization for the second-stage optimization to ensure more reliable and high-quality policy discovery.
\vspace{-10pt}

\vspace{-15pt}
\section{Quality-Efficiency Trade-off}
\label{sec:trade-off}

\begin{figure}[h]
    \centering
    \includegraphics[width=1.0\linewidth]{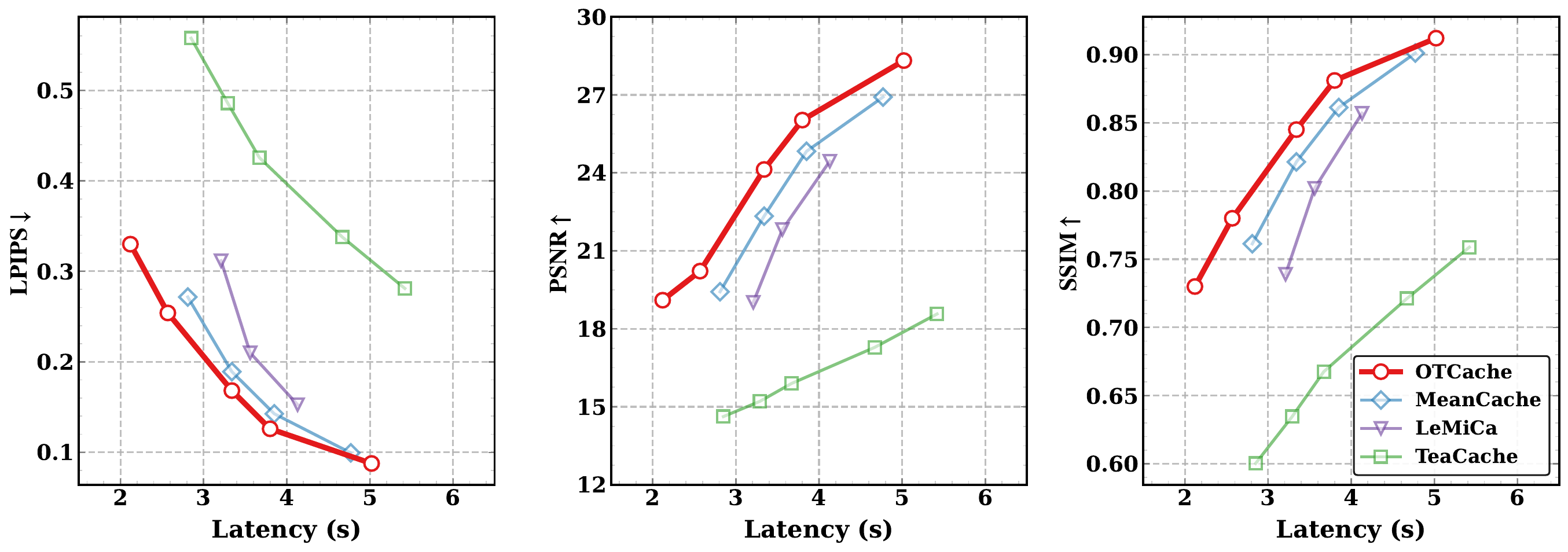}
    \caption{Quality-latency comparison across different caching methods.}
    \label{fig:trade_off}
\end{figure}

We evaluate the quality-efficiency trade-off across varying inference budgets $\mathcal{B} \in \{20, 15, 13, 10, 8\}$, representing a latency range of 2-5 seconds. As shown in Fig.~\ref{fig:trade_off}, OTCache consistently establishes a superior Pareto front compared to state-of-the-art baselines including TeaCache, LeMiCa, and MeanCache. Notably, OTCache maintains lower LPIPS and higher PSNR/SSIM across all latency configurations. The performance gap becomes more pronounced as the budget decreases, demonstrating the robustness of our Optimal Transport-based interpolation in identifying high-quality caching schedules under tight computational constraints.

\vspace{-15pt}

\section{Offline Calibration Cost}
\label{sec:stage2_cost}

Stage-2 in OTCache is not executed online for each new user prompt. Instead,
it is an offline one-time calibration procedure for a given backbone model and
anchor budget. In our experiments, we construct the calibration set by sampling
50 prompts from T2V-CompBench, with no overlap with the evaluation prompts.
We then optimize Eq~\eqref{eq:stage2_e2e_obj} under $B_{\mathrm{anc}}=8$ using Optuna with a budget
of 200 trials and an early-stop patience of 50. The resulting anchor schedule is
used as an offline calibrated policy for all subsequent inference under the same
model setting.

\vspace{-15pt}

\begin{table}[h]
\centering
\caption{Per-prompt offline calibration cost of Stage-2. The reported worst-case
cost assumes that all 200 trials are executed. In practice, early stopping can further
reduce the calibration overhead.}
\vspace{-2mm}
\label{tab:stage2_cost}
\setlength{\tabcolsep}{4pt}
\footnotesize
\resizebox{\linewidth}{!}{%
\begin{tabular}{lccccc}
\toprule
Model & Anchor NFE & Trials & Early Stop & Per-trial Cost & Worst-case Cost \\
\midrule
FLUX.1 & 8 & 200 & 50 & 2.3 s & 7.7 min \\
Qwen-Image & 8 & 200 & 50 & 5.5 s & 18.2 min \\
HunyuanVideo & 8 & 200 & 50 & 15.1 s & 50.5 min \\
\bottomrule
\end{tabular}}
\end{table}

\vspace{-15pt}

During online inference, OTCache directly uses the calibrated anchor and
therefore does not incur the search cost of Stage-2. The only additional online
cost is the negligible schedule prediction step in Stage-3, which consists of
interpolation and discrete schedule realization. Table~\ref{tab:stage2_cost} reports the worst-case per-prompt calibration cost
when all 200 trials are executed. In practice, early stopping often reduces this
cost. Since calibration prompts are independent, the offline search naturally
supports multi-GPU data parallelism. On 8$\times$H100 GPUs, calibrating the full
50-prompt set is estimated to take about 39 min, 1.5 h, and 4.2 h for FLUX.1,
Qwen-Image, and HunyuanVideo, respectively.

\section{More Visual Comparison}
\label{sec:more_visual}

\begin{figure*}[h]
  \centering
  \vspace{-2pt}
  \includegraphics[width=0.95\linewidth]{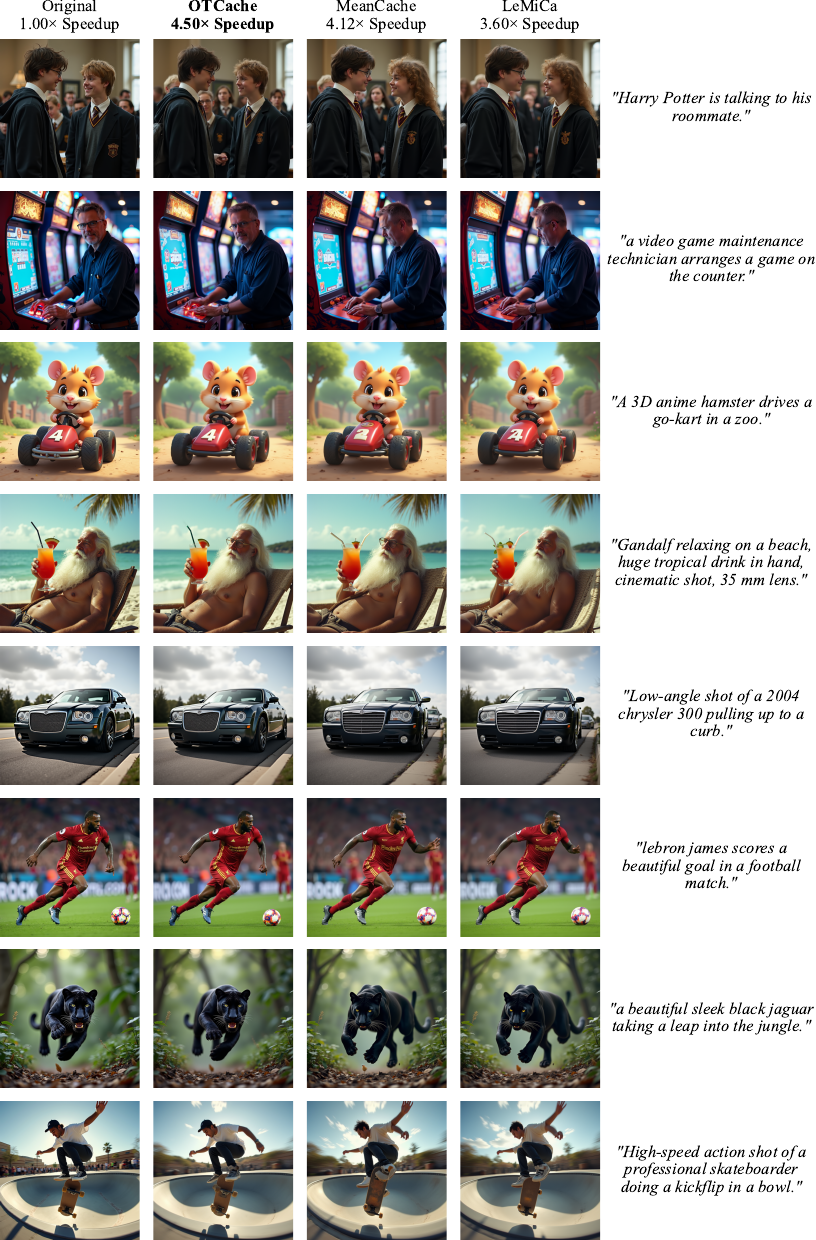}
  \vspace{-2pt}
  \caption{More visual comparisons on \textbf{FLUX.1 [dev]}, Best viewed zoomed in.}
  \vspace{-5mm}
  \label{fig:appendix_flux}
\end{figure*}

\begin{figure*}[h]
  \centering
  \vspace{-2pt}
  \includegraphics[width=0.95\linewidth]{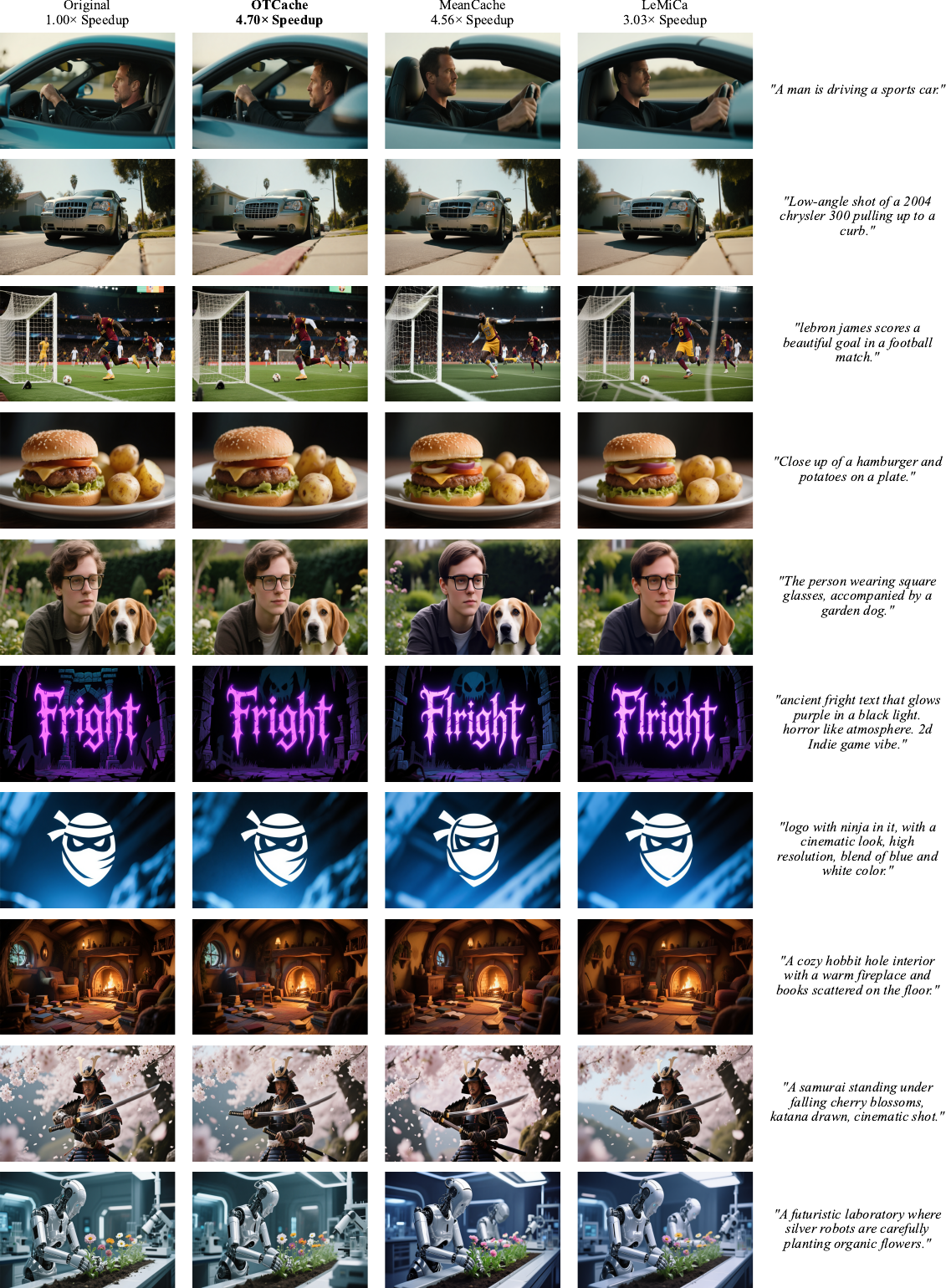}
  \vspace{-2pt}
  \caption{More visual comparisons on \textbf{Qwen-Image}, Best viewed zoomed in.}
  \vspace{-5mm}
  \label{fig:appendix_qwen}
\end{figure*}

\begin{figure*}[h]
  \centering
  \vspace{-2pt}
  \includegraphics[width=0.95\linewidth]{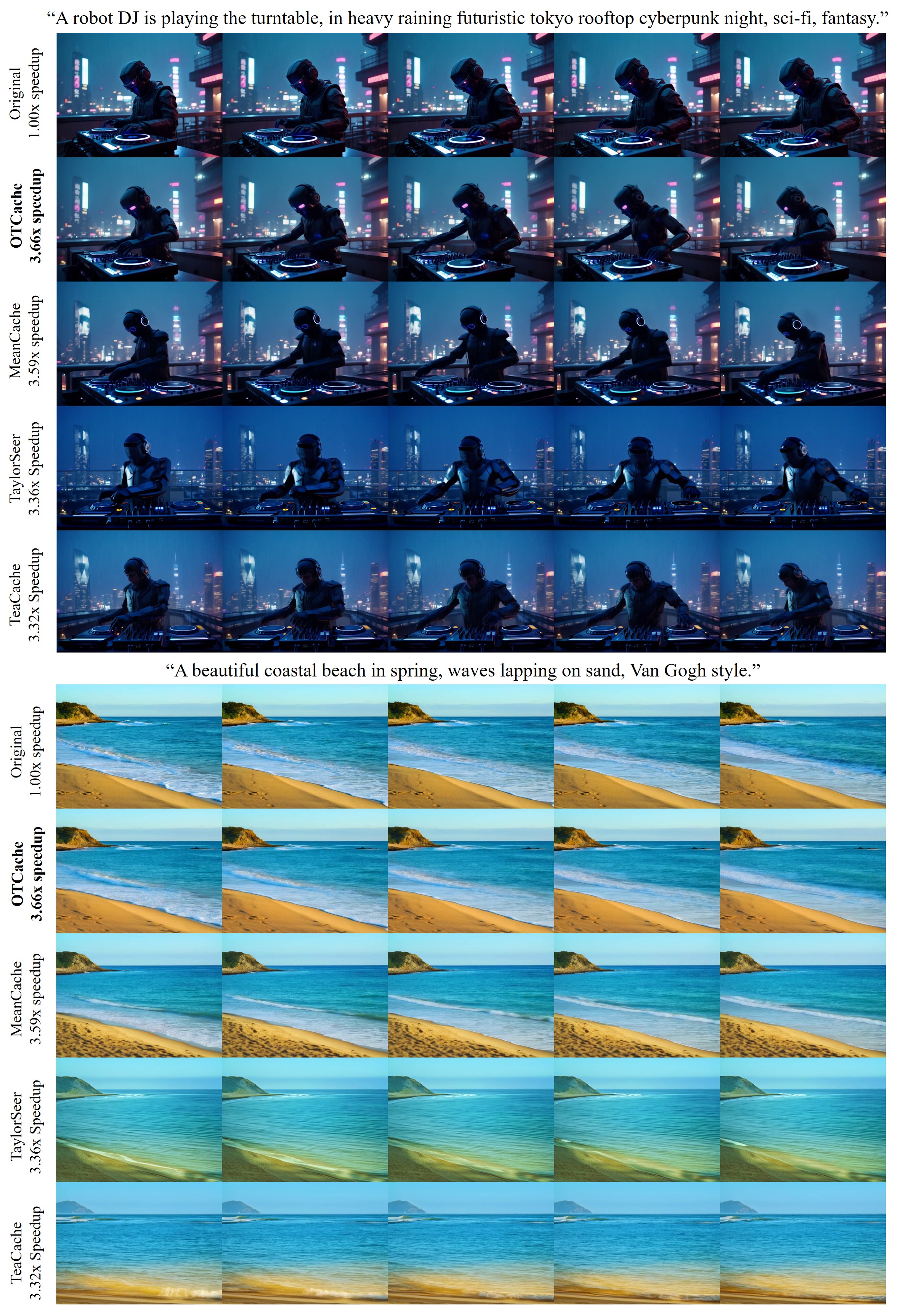}
  \vspace{-2pt}
  \caption{More visual comparisons on \textbf{HunyuanVideo} (1/3), Best viewed zoomed in.}
  \vspace{-5mm}
  \label{fig:compare_video_1}
\end{figure*}

\begin{figure*}[h]
  \centering
  \vspace{-2pt}
  \includegraphics[width=0.95\linewidth]{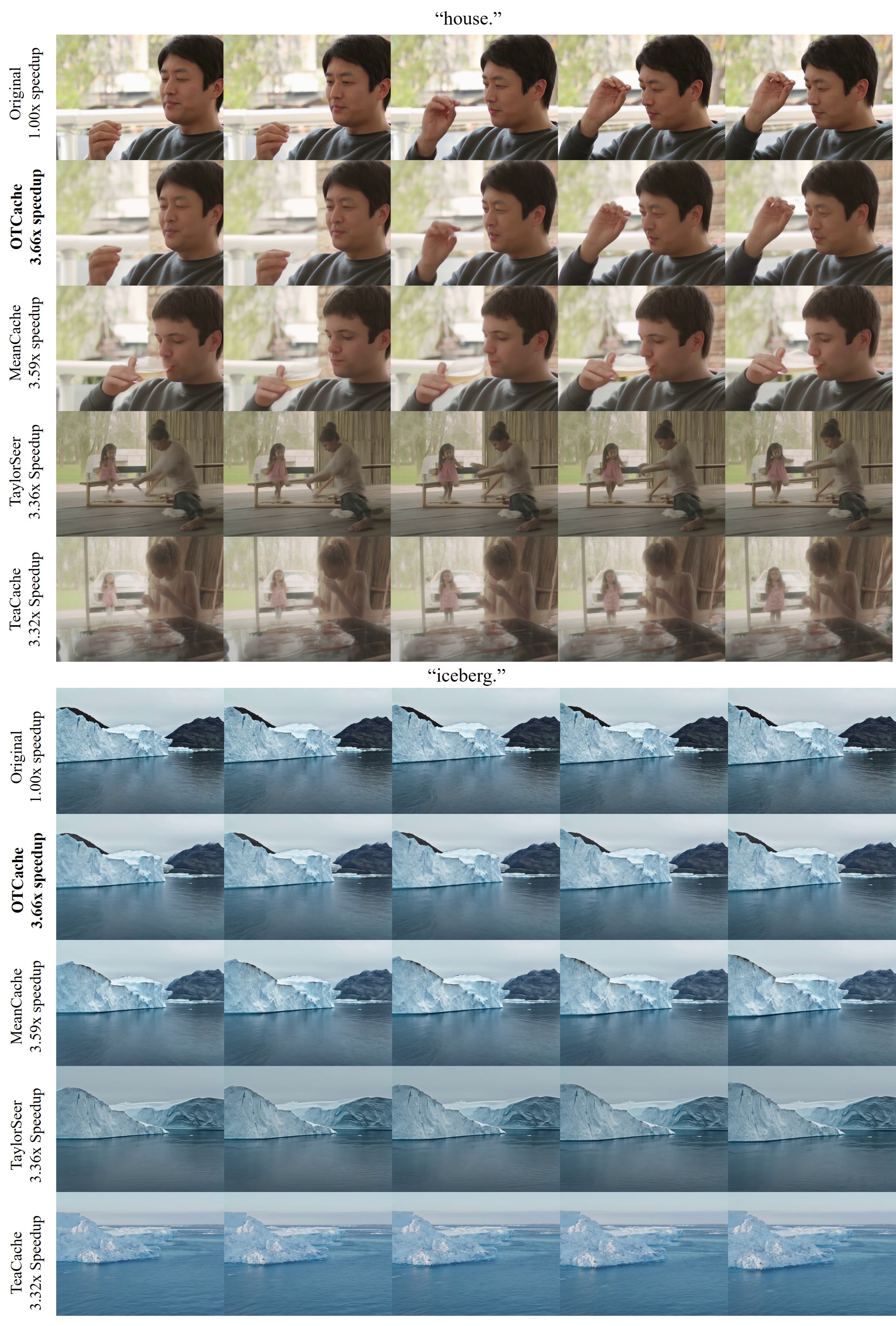}
  \vspace{-2pt}
  \caption{More visual comparisons on \textbf{HunyuanVideo} (2/3), Best viewed zoomed in.}
  \vspace{-5mm}
  \label{fig:compare_video_2}
\end{figure*}

\begin{figure*}[h]
  \centering
  \vspace{-2pt}
  \includegraphics[width=0.95\linewidth]{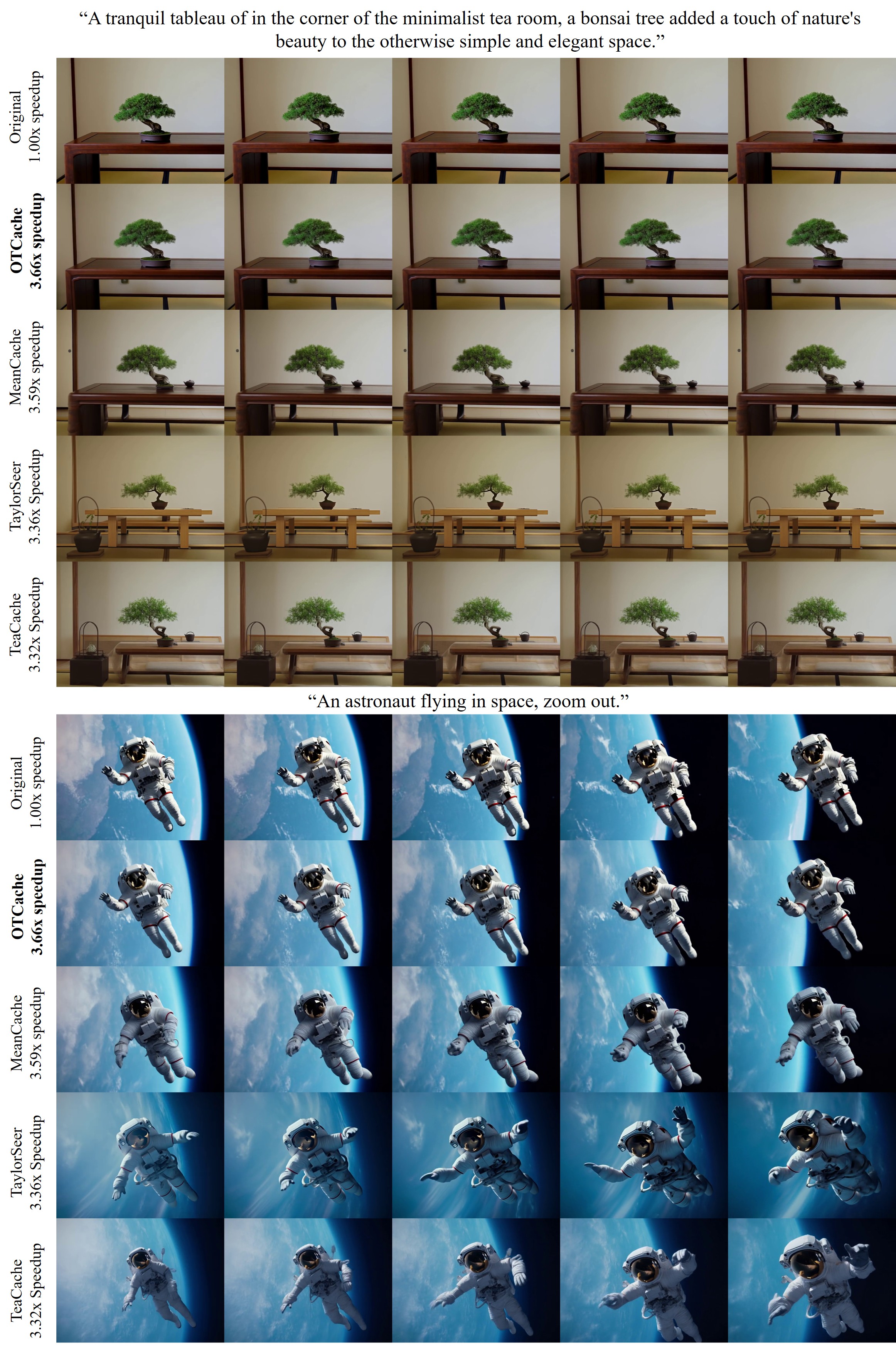}
  \vspace{-2pt}
  \caption{More visual comparisons on \textbf{HunyuanVideo} (3/3), Best viewed zoomed in.}
  \vspace{-5mm}
  \label{fig:compare_video_3}
\end{figure*}

To further evaluate the robustness of \textbf{OTCache}, we provide extensive qualitative comparisons across three representative architectures: FLUX.1 [dev] (Fig.~\ref{fig:appendix_flux}), Qwen-Image(Fig.~\ref{fig:appendix_qwen}), and HunyuanVideo(Fig.~\ref{fig:compare_video_1}-\ref{fig:compare_video_3}). All samples are generated under the maximum acceleration ratios (Low-NFE) to ensure a rigorous assessment. As illustrated, OTCache consistently preserves superior structural integrity and fine-grained textures in images while effectively eliminating flickering in videos. Even under extreme constraints, our method maintains high-fidelity content and temporal coherence by following the smooth evolution of caching policies, significantly outperforming baselines.

\end{document}